\documentclass[Afour,sageh,times]{sagej}

\usepackage[T1]{fontenc}
\usepackage{lmodern}   

\usepackage{moreverb,url}
\usepackage{amssymb}
\usepackage{amsmath}

\usepackage{booktabs}
\usepackage{subcaption}
\usepackage{graphicx}
\usepackage{enumitem}
\usepackage{xcolor}
\usepackage{multirow}
\usepackage[linesnumbered,ruled,vlined]{algorithm2e}
\usepackage{algpseudocode}
\usepackage{dblfloatfix}
\usepackage{amsthm}
\usepackage{titlesec}

\titleformat{\subsubsection}[block] 
  {\normalfont\normalsize\bfseries} 
  {\thesubsubsection}{1em}{}       

\usepackage[colorlinks,bookmarksopen,bookmarksnumbered,citecolor=blue,urlcolor=blue,linkcolor=blue]{hyperref}

\newcommand\BibTeX{{\rmfamily B\kern-.05em \textsc{i\kern-.025em b}\kern-.08em
T\kern-.1667em\lower.7ex\hbox{E}\kern-.125emX}}

\newcommand{\methodname}{SLAC}

\DeclareMathOperator*{\bS}{\mathcal{S}}

\DeclareMathOperator*{\bZ}{\mathcal{Z}}

\begin{document}

\runninghead{Hu et al.}

\title{\methodname{}: Safe and Efficient Real-Robot Reinforcement Learning via Unsupervised Simulation Pre-Training}


\author{Jiaheng Hu\affilnum{1}, Peter Stone\affilnum{1,2}, Roberto Mart\'{i}n-Mart\'{i}n\affilnum{1,3}}

\affiliation{\affilnum{1}The University of Texas at Austin\\
\affilnum{2}Sony AI\\
\affilnum{3}Amazon Robotics}

\corrauth{Jiaheng Hu, The University of Texas at Austin,
USA.}

\email{jiahengh@utexas.edu \\}

\begin{abstract}
Robotic Reinforcement Learning faces a persistent dilemma: accurate simulators can enable safe and scalable training, but are expensive to build and often fail to capture the contact-rich dynamics needed for real-world deployment; direct real-world learning avoids simulator mismatch, but exploration in high-dimensional robot action spaces is slow and unsafe. We propose SLAC, a framework that resolves this tension by using low-fidelity simulation as an affordance-discovery substrate rather than as a source of transferable task policies. SLAC first learns a task-agnostic latent action decoder in a coarse simulator that preserves approximate geometric and interaction affordances while omitting task-specific rewards and high-fidelity dynamics. The decoder is trained with a disentangled unsupervised skill discovery objective and a safety-aware exploration reward, yielding temporally extended latent actions that provide a structured interface for real-world learning. SLAC then trains downstream policies directly in the real world, via a novel off-policy RL method that exploits the learned action factorization through masked Q-decomposition and autonomous action-reward dependency detection. In real-world experiments with a bimanual mobile manipulator, SLAC learns contact-rich whole-body tasks in less than an hour of autonomous real-world interaction, without relying on any demonstrations or hand-crafted behavior priors. These results substantially outperform prior methods, and establish SLAC as a promising framework for safe and efficient robot RL.
More information, code, and robot videos at \url{robo-rl.github.io}.
\end{abstract}



\maketitle


\section{Introduction}
\label{s:intro}

\begin{figure*}[t]
\centering
\includegraphics[width=0.92\textwidth, trim=110 80 300 130, clip]{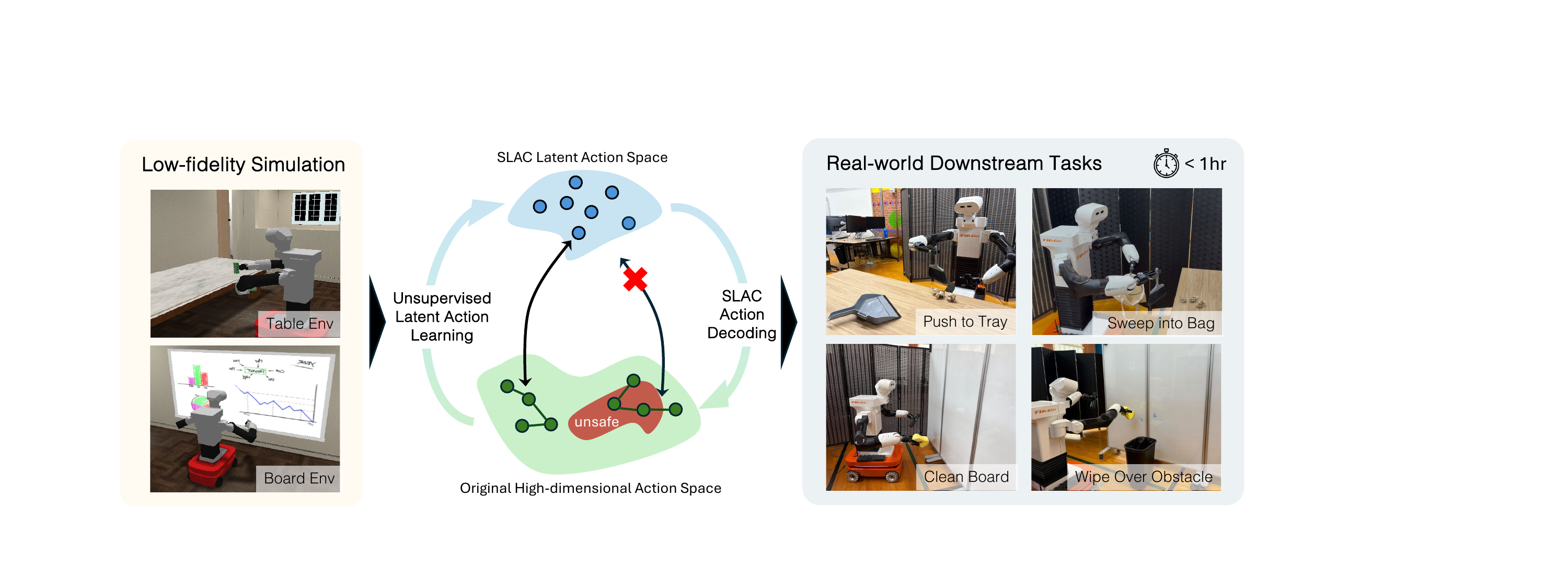}
\caption{\methodname{} uses a task-agnostic action space trained in low-fidelity simulation (\textit{left}) to learn downstream tasks in the real world. This latent action space is safe, temporally extended, and disentangled, enabling a bimanual mobile manipulator to solve challenging contact-rich whole-body tasks (\textit{right}) with less than an hour of autonomous real-world interactions.
}
\label{fig:env_viz}
\end{figure*}

Recently, Reinforcement Learning (RL) has achieved remarkable successes in Robotics~\citep{tang2024deep}, spanning manipulation~\citep{andrychowicz2020learning}, locomotion~\citep{lee2020learning}, drone racing~\citep{kaufmann2023champion}, and robot  athletes~\citep{durr2026outplaying, haarnoja2024learning}. Most of these successes share a similar recipe: conduct RL in a high-fidelity simulation that matches the real world as closely as possible, then deploy the learned policy zero-shot on a real robot.
Such a sim-to-real pipeline conveniently mitigates two of the major challenges in RL. On the one hand, the high \textbf{sample complexity} of widely used RL algorithms (e.g., PPO~\citep{schulman2017proximal}) is alleviated by the ability of modern simulators to generate a large amount of rollouts in parallel. On the other hand, \textbf{safety concerns} associated with robots damaging themselves during random exploration are eliminated, as training is conducted entirely in simulation.

Unfortunately, zero-shot sim-to-real transfer places extremely high requirements on having accurate simulation environments.
These simulation environments require time-consuming creation of simulation assets (e.g. robot URDF, object meshes), and iterative system identification to match their physical properties. Such a process can take months of iterative tuning~\citep{li2025robotmover,tang2024deep,fu2023deep,he2025asap,zhao2020sim}, and often has to be repeated separately for every new scenario or task.
Furthermore, despite numerous attempts~\citep{jiang2025phystwin,si2024difftactile,li2023behavior}, certain classes of physical interactions, such as deformable objects (e.g., sponges), contact-induced changes (e.g., marker traces), and fluid behaviors (e.g., pouring water), remain extremely challenging to simulate accurately with existing techniques, resulting in brittle policies that collapse due to the sim-to-real gap.
Together, these drawbacks significantly limit the scalability and general applicability of zero-shot sim-to-real transfer.

Can we leverage an \emph{inexpensive, low-fidelity simulator} that is easy to create, but still benefit from the advantages in \emph{sample complexity and safety} brought by sim-to-real RL?
Our key insight is that a low-fidelity
simulator may be insufficient for transferring a task policy,
but can still be sufficient for learning reusable action
abstractions. Even when the simulator does not reproduce
the visual appearance, contact dynamics, or downstream
task reward of the real world, it can expose affordance-level
structure: which surfaces can be reached, which contacts
are possible, and which regions are unsafe. 
By learning such action abstractions
in simulation and composing them during downstream
learning in the \textbf{real world}, we can improve both \textbf{sample
efficiency and exploration safety} without requiring access to
a high-fidelity simulator.

This article introduces \textbf{\methodname{}}: \textbf{S}imulation-Pretrained \textbf{L}atent \textbf{AC}tion Space for Real-World RL, a framework which utilizes \textit{low-fidelity simulators} to make safe and efficient real-world downstream RL feasible for challenging robotics problems. 
The \methodname{} framework consists of a two-step procedure. 
In the first step, \methodname{} learns a task-agnostic latent action space in a coarsely aligned, low-fidelity simulator via Unsupervised Skill Discovery (USD)~\citep{eysenbach2018diversity,wang2024skild,park2023controllabilityawareunsupervisedskilldiscovery,hu2024disentangled}. \methodname{} utilizes a novel USD objective that shapes this latent action space to be (1) \textbf{temporally extended}, enabling effective exploration by reducing decision frequency; (2) \textbf{disentangled}, allowing each latent action dimension to independently affect the states, thereby facilitating joint optimization of multiple objectives without conflict; and (3) \textbf{safe}, avoiding dangerous behaviors that could damage the robot. In the second step, the learned \methodname{} latent action space is used by a novel off-policy RL algorithm to efficiently learn downstream tasks directly in the real world. Critically, this design offers robustness to the sim-to-real gap: even if latent actions exhibit slight behavioral mismatches between simulation and the real world, the downstream policy can still learn to solve the task by directly selecting effective latent actions based on real-world reward signals.

Empirically, \methodname{} enables a bimanual mobile manipulator to learn contact-rich whole-body tasks in less than an hour of real-world interactions, using only onboard sensor signals. To the best of our knowledge, \methodname{} is the first algorithm that enables a high-DoF mobile manipulator to learn with RL in the real world without relying on any demonstrations/mocap data~\citep{xiong2024adaptive, herzog2023deep} or hand-crafted behavior priors~\citep{xiong2024adaptive, sun2022fully, mendonca2024continuously}.

In summary, this work makes the following main contributions:
\begin{itemize}[itemsep=-5pt, topsep=-1pt, leftmargin=*]
    \item A framework for real-robot RL based on a task-agnostic latent action space learned in low-fidelity simulation.
    \item An unsupervised RL algorithm for learning disentangled and safety-aware latent action space that are suitable for downstream learning. 
    \item An off-policy RL algorithm that autonomously inferences and leverages the factorized structure of the learned latent action space for sample-efficient downstream task learning in the real world.
    \item Empirical validation of our framework, including learning challenging contact-rich whole-body visuomotor tasks on a real tiago robot without any demonstration or hand-crafted behavior priors.
\end{itemize}

\paragraph{Novelty Statement:}
An earlier version of this article was presented at the Conference on Robot Learning~\citep{hu2025slac}. This article significantly extends upon the conference version in three major aspects:


\begin{enumerate}[itemsep=-5pt, topsep=-1pt,leftmargin=*]
    \item We generalize the \methodname{} framework beyond the original mobile manipulation setting (Sec.~\ref{s:intro}, Sec.~\ref{s:formulation}), with novel experimental results in multi-robot domains (Sec.~\ref{ss:par}, Sec.~\ref{ss:abla}).
    \item We introduce a method for accelerating latent action space learning via Q-decomposition (Sec.~\ref{ss:dec}). In Sec.~\ref{ss:abla}, we show experimental results to demonstrate the effectiveness of this technique.
    \item For downstream task learning, we introduce a pipeline for autonomous detection of the dependencies between the latent action dimensions and the reward terms, as elaborated on in Sec.~\ref{ss:ar_dep}. This improvement enables the downstream learning pipeline to be fully autonomous.
\end{enumerate}



\section{SLAC Problem Formulation}
\label{s:formulation}


\methodname{} aims to enable sample-efficient and safe real-world reinforcement learning (RL) for challenging robotics tasks. We formulate the real-world RL problem as a Partially Observable Markov Decision Process (POMDP), defined by the tuple $\mathcal{M} = (\mathcal{S}, \mathcal{A}, \mathcal{O}, P, R_{\mathit{task}}, \gamma)$, where $\mathcal{S}$ is the set of underlying environment states, $\mathcal{A}$ is the high-dimensional native action space (e.g., joint velocities or torques), $\mathcal{O}$ is the observation space (e.g. camera images), $P(s'|s, a)$ is the state transition function, $R_\mathit{task}(s, a) = \sum_{i=1}^m R_i(s,a)$ is a composite reward function with $m \geq 1$ term(s)\footnote{This formulation is general, as any reward function can be expressed as a sum of component functions.}, and $\gamma \in (0, 1]$ is the discount factor. The objective is to learn a policy $\pi(a|o)$ that maximizes the expected return:
\begin{equation}
    \pi^* = \arg\max_\pi \mathbb{E}_{\pi} \left[ \sum_{t=0}^\infty \gamma^t R_{\mathit{task}}(s_t, a_t) \right]
\end{equation}
Due to the high dimensionality of $\mathcal{A}$ and the complexity of real-world tasks, directly optimizing $\pi(a|o)$ in the real world is prohibitively sample-inefficient and unsafe.
To address these issues, we propose to replace the native control space $\mathcal{A}$ with a \textit{latent action space} $\mathcal{Z}$ learned in a \textbf{low-fidelity simulation}. The simulation does not accurately replicate the visual or physical properties of the real world and does not implement the task reward $R_\mathit{task}$, but approximately retains key physical affordances and shares the same robot action space $\mathcal{A}$.

Specifically, we aim to learn a \textbf{latent action decoder} $\pi_{\mathit{dec}}(a | o_{\mathit{dec}}, z)$, which converts a latent action $z \in \mathcal{Z}$ into low-level actions $a \in \mathcal{A}$ based on a low-dimensional decoder observation $o_{\mathit{dec}}$ that is shared across simulation and the real world (e.g., proprioceptive states, furniture poses). We discuss in Sec.~\ref{ss:la} how we learn this latent action decoder through a novel unsupervised skill discovery algorithm.

Once the latent action decoder is learned, \methodname{} trains a perception-to-latent \textbf{task policy} $\pi_{\mathit{task}}(z | o)$ in the real world given a downstream task reward.
$\pi_{\mathit{task}}(z | o)$ selects latent actions based on (history of) high-dimensional real-world observations $o \in \mathcal{O}$ (e.g. camera images), and is \textbf{trained entirely in the real world} using a novel sample-efficient off-policy RL method explained in Sec.~\ref{ss:ds}.

Together, the task policy and the latent action decoder define a hierarchical visuomotor policy over low-level robot actions, which can be run directly on a real robot with on-board sensors:
\begin{equation}
    \pi(a | o) = \int_z \pi_{\mathit{dec}}(a | o_{\mathit{dec}}, z)\, \pi_{\mathit{task}}(z | o)\, dz
\end{equation}

\begin{figure*}[t]
  \centering
  \includegraphics[width=0.95\textwidth, trim=75 120 170 110, clip]{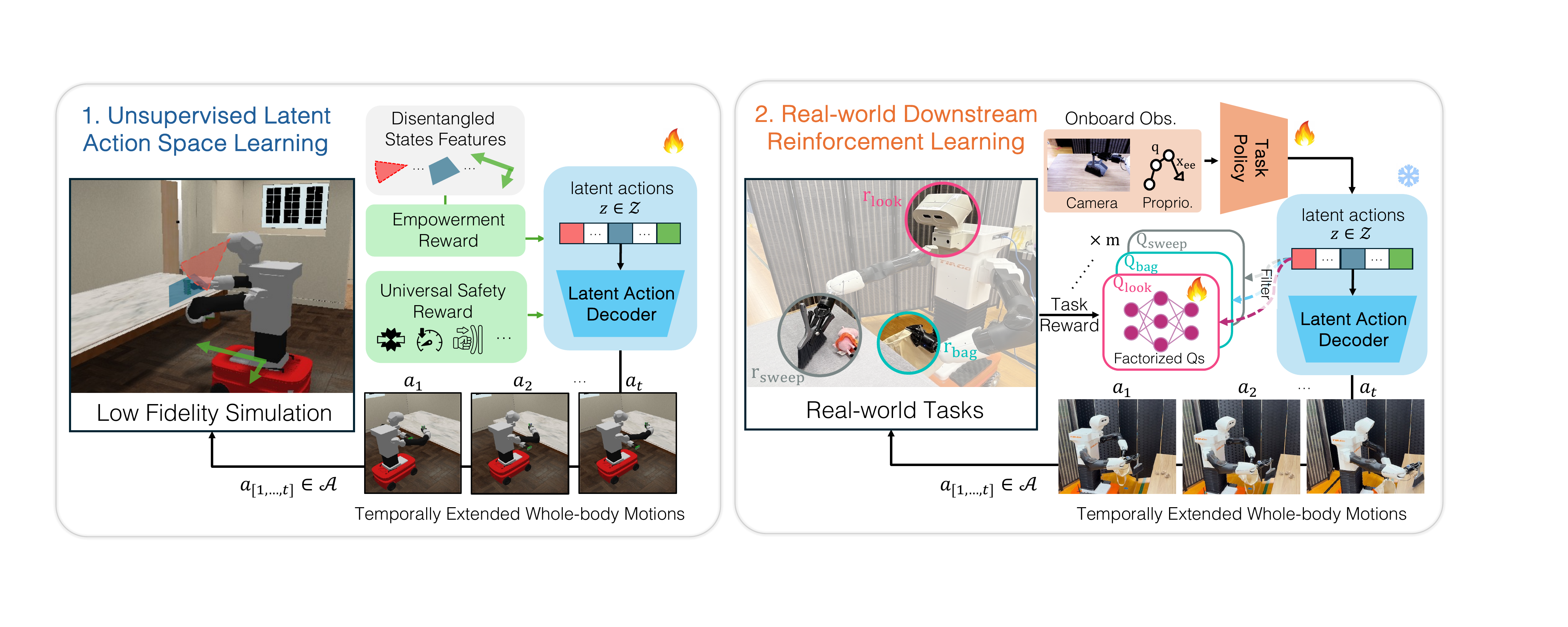}
  \caption{The two-step \methodname{} procedure to enable real-world policy learning. (\textit{Left}) In the first step, \methodname{} learns a \textbf{Latent Action Decoder} that maps each latent action, $z\in\mathcal{Z}$, to a sequence of low-level robot actions, $(a_0,\ldots,a_T), a_t\in\mathcal{A}$. This decoder is learned in low-fidelity simulation via unsupervised skill discovery with novel objectives that encourage the robot to independently control different state features (e.g., \textcolor{red}{camera directions}, \textcolor{blue}{contacts with table}, \textcolor{green}{base locations}) while being safe. (\textit{Right}) In the second step, once the decoder is trained, the robot learns downstream tasks with RL in the real world using the SLAC latent action space. The task policy directly takes in the onboard sensor observations of the robot (i.e., images, proprioception) and outputs latent actions $z$ that are decoded into safe robot actions. \methodname{} applies Factorized Latent-Action SAC to optimize the policy for downstream tasks with multi-term reward (e.g., \textcolor{magenta}{look at the objects}, \textcolor{teal}{keep a bag close}, \textcolor{gray}{sweep the trash}) directly in the real world with very few samples, converging in less than an hour, by taking advantage of high-frequency off-policy updates and masked Q decomposition.}
  \label{fig:pipeline}
\end{figure*}

We show the full pipeline of our method in Fig.~\ref{fig:pipeline}

\section{\methodname{} Step 1: Learning Latent Action Space in Simulation}
\label{ss:la}


The first step of \methodname{} leverages simulation to learn a task-agnostic latent action space capable of supporting a wide range of real-world task variations.
\methodname{} acquires such an action space via the framework of Unsupervised Skill Discovery (USD)~\citep{eysenbach2018diversity,wang2024skild,park2023controllabilityawareunsupervisedskilldiscovery}, which allows agents to learn diverse, temporally-extended, task-agnostic behaviors purely from interactions without relying on explicit task rewards. 
This process yields a latent skill decoder $\pi_\mathit{dec}(a|o_\mathit{dec},z)$, where each latent skill $z$ induces a distinct behavior.
These learned skills can then be composed by a task policy $\pi_\mathit{task}(z|o)$ to efficiently solve downstream tasks, where the learned skill space serves as a temporally extended action space of the task policy.\footnote{For the rest of this paper, we will use ``skills'' and ``latent actions'' interchangeably.}


\methodname{} conducts unsupervised skill discovery entirely in simulation, where data collection is fast and inexpensive. Our key insight is that even \textbf{low-fidelity simulation} can serve as an effective substrate for behavior pretraining: \emph{as long as the learned skills span a sufficiently diverse range of behaviors, they can be composed downstream to solve real-world tasks.}
To this end, \methodname{} employs simulation environments that may not directly replicate a real-world counterpart (e.g. not visually realistic, no hard-to-simulate objects like marker traces, no reward), but still preserve key geometric affordances that are potentially useful for downstream tasks (e.g. a whiteboard that the robot can touch, an obstacle that the robot may collide with). We show some of these environments in Fig.~\ref{fig:env_viz}.
Importantly, because we do not need the simulation environment to exactly match the real world, prototyping a new environment and training a corresponding latent action space can be done quickly.


In Sec.~\ref{ss:dusdi}, we introduce a novel USD algorithm that facilitates efficient downstream task learning through creating a structured latent action space. In Sec.~\ref{ss:safe}, we introduce a method to ensure that the learned latent action space is safe. In Sec.~\ref{ss:dec}, we introduce a mechanism that leverages Q-function decomposition to accelerate and stabilize the learning of latent action space.
We show the pseudo-code for latent action space learning in Algo.~\ref{app:alg}, and further discuss the properties of the learned action space in App.~\ref{app:la}.

\subsection{Disentangled Unsupervised Skill Discovery for Structured Latent Action Space}

\label{ss:dusdi}


Many unsupervised RL methods have been proposed in the past~\citep{eysenbach2018diversity,park2023controllabilityawareunsupervisedskilldiscovery,gregor2016variational,laskin2021urlb, wang2024skild, sharma2019dynamics}, with impressive results on learning diverse behaviors without any task reward.
However, when it comes to using these learned behaviors as a latent action space for solving downstream tasks, these methods often fall short.

Our main realization is that this performance deficiency arises from an unstructured latent action space (e.g. a one-hot vector as in \cite{eysenbach2018diversity}), in which action representations lack semantically meaningful internal structure. This lack of structure causes the effect of the latent action to be \textbf{entangled}, where 
taking a step in the latent action space leads to arbitrary influences on \textit{multiple dimensions} of the state space simultaneously.
Learning to use and recombine these \textit{entangled} latent actions can therefore be extremely hard for an agent trying to solve downstream tasks.

To tackle this issue, we introduce a novel USD algorithm, Disentangled Unsupervised Skill Discovery (DUSDi)\footnote{DUSDi (Sec.~\ref{ss:dusdi}) was initially presented in our prior work \citep{hu2024disentangled} for Hierarchical RL of virtual agents.}, which defines a factored latent action space, and encourages \textbf{disentanglement} between different latent action factors through a novel learning objective. Such a latent action space creates a clear mapping between latent action factors and interactable environment entities, enabling efficient downstream learning, especially for \emph{high degree-of-freedom, multi-purpose robots}.\\


\subsubsection{DUSDi Objective}

DUSDi introduces a novel disentanglement objective grounded in a factored state space assumption, where $\{\mathbf{S}^i\}_{i=1}^N$ represents a set of state entities (e.g. whiteboard, table, robot position) in the environment that the robot agent can influence. Given these state factors, which are often naturally available in simulation and/or easy to define manually, DUSDi defines a factorized latent action space 
$\mathcal{Z} = \bZ^1 \times \dots \times \bZ^N$ to match the structure of the factored states. 
We define each $\bZ^i$ as a discrete categorical variable with $k$ categories\footnote{We discuss extension to continuous latent actions in App.~\ref{app:discrete}}. Each latent action is therefore an $N$ dimensional multi-categorical vector with $k^N$ possible values.



Given this factored latent action space, our goal is to learn a latent action decoder, $\pi_\mathit{dec}(a|o_\mathit{dec},z)$, such that each latent action factor ${\bZ}^i$ affects and only affects (through the action decoder) the value of a particular state factor, ${\bS}^i$. 
For each latent action factor and state factor pair $(\bZ^i, \bS^i)$, we encourage diverse and distinguishable behaviors by maximizing their mutual information $I({\bS}^i;{\bZ}^i)$\footnote{See \cite{eysenbach2018diversity} for more detailed justification of the mutual information objective.}.
While this objective enables each latent action factor to affect the corresponding state factor, it does not restrict the action factor from affecting other state factors.
As a result, the latent action factors would still be entangled in their effects. To prevent this issue, we propose to ensure that each latent action factor, ${\bZ}^i$, minimally affects the rest of the state factors, ${\bS}^{\neg i}$, where ${\bS}^{\neg i}$ denotes the subspace formed by all other state factor spaces except ${\bS}^{i}$: $\bS^1\times \dots, \bS^{i-1}\times \bS^{i+1}\times \dots\times \bS^N$.
Specifically, we incorporate an entanglement penalty to minimize, $I({\bS}^{\neg i}; {\bZ}^i)$, which corresponds to the mutual information between a latent action factor and all other state factors that it should not affect.

Formally, the latent action decoder aims to maximize the following objective:
\begin{equation}
\mathcal{J}(\theta) = \sum_{i=1}^N I({\bS}^i;{\bZ}^i) - \lambda I({\bS}^{\neg i}; {\bZ}^i), \label{eq:obj}
\end{equation}


where $\lambda \in [0, 1)$ is a hyperparameter that controls the importance of the entanglement penalty relative to the action-state association. We recommend using a small $\lambda$ (e.g. $\lambda=0.1$) for the following reason: in some environments, due to intrinsic dynamical dependencies between state factors themselves, controlling a state factor, $\bS^i$, has to introduce some association between $\bZ^i$ and other factors in $\bS^{\neg i}$, e.g., when controlling an object whose manipulation requires the agent to use other objects as tools.
In these cases, as the action decoder learns to maximize the MI between an action factor and a state factor, $I(\bS^i, \bZ^i)$, the MI with other state factors, $I({\bS}^{\neg i}; {\bZ}^i)$, may also increase.
For these cases, the use of $\lambda < 1$ will ensure that the entanglement penalty does not overpower the association reward, and the policy is still incentivized to learn $\bZ^i$ that change $S^i$ distinguishably while introducing minimal changes on other state factors. \\

\begin{algorithm*}[ht]
\caption{\methodname{} Unsupervised Latent Action Space Learning\\
\smallskip
\textit{\hspace{2em}Note: Omitting standard SAC steps (e.g., target network creation and updates) for simplicity.}
}
\label{app:alg}

Create $\mathit{sim}$ environment, skill prior distribution $p(z)$, replay buffer $\mathcal{D}_\mathit{sk}$\;
Initialize latent action decoder $\pi_\mathit{dec}$, N discriminators $q^i_\phi$, $q^i_\psi$, and N value functions $Q^i_\mathit{dec}$\;
\For{$k \leftarrow 1$ \KwTo $skill\_learning\_epochs$}{
    Sample latent action $z \sim p(z)$\;
    \For{$j \leftarrow 1$ \KwTo $steps\_per\_skill$}{
        $(o_\mathit{dec}, a,{o'}_\mathit{dec}) \gets \texttt{sim.step}(\pi_\mathit{dec}(a \mid o_\mathit{dec}, z))$\;
        Store transition $(o_\mathit{dec}, a, z, {o'}_\mathit{dec})$ into replay buffer $\mathcal{D}_\mathit{sk}$\;
        \For{$i \leftarrow 1$ \KwTo $n\_updates$}{
            Sample mini-batch $\{(o_\mathit{dec}, a, z, {o'}_\mathit{dec}\}$ from $\mathcal{D}_\mathit{sk}$\;
            Update $q_\phi$ and $q_\psi$ via gradient ascent\;
            Calculate intrinsic reward $r$ based on Eq.~\ref{eq:final}\;
            Update $Q_\mathit{dec}$ with $r$ using SAC critic update and  Q-Decomposition\;
            Update $\pi_\mathit{dec}$ with $Q_\mathit{dec}$ using SAC policy update\;
        }
    }
}
\KwRet{$\pi_\mathit{dec}$}
\end{algorithm*}

\subsubsection{DUSDi Tractable Optimization}

Directly maximizing the objective in Eq.~\ref{eq:obj} is intractable. 
Instead, we propose to approximate the objective using a variational lower bound of the mutual information~\citep{barber2003information}:
\begin{align}
I({\bS}^i;{\bZ}^i) = H({\bZ}^i) - H({\bZ}^i|{\bS}^i) \geq C +  \mathbb{E}_{z, s} \log {q^i_\phi} (z^i| s^i),  \label{eq:MI1}
\end{align}
where $C$ represents the constant value of $H({\bZ}^i)$, the entropy of the prior distribution over the latent action variable, which does not change during training, and $q^i_\phi$ is a variational distribution (e.g. a neural network discriminator mapping input state factor to the predicted disentangled component value, $z^i$).

Similarly, we can approximate the MI in the entanglement penalty by:
\begin{align}
I({\bS}^{\neg i};{\bZ}^i) \geq C +  \mathbb{E}_{z, s} \log {q^i_\psi} (z^i| s^{\neg i}), \label{eq:MI2}
\end{align}
where $q^i_\psi$ is another variational distribution.
Critically, when these $q$ approximations perfectly recover the posterior distribution of $z^i$, we obtain equality in Eq.~\ref{eq:MI1} and Eq.~\ref{eq:MI2}, and hence recover the true objective in Eq.~\ref{eq:obj}. Therefore, to optimize $\mathcal{J}(\theta)$, we alternate between two steps: 1) performing variational inference to train the discriminators $q^i_\phi$ and $q^i_\psi$ through gradient ascent,
and 
2) using $q^i_\phi$ and $q^i_\psi$ to learn a latent action decoder $\pi_\mathit{dec}$ through RL by maximizing the following intrinsic reward approximating Eq.~\ref{eq:obj}:
\begin{equation} \label{eq:intri_rew}
r_\mathit{dusdi}(s, a) \triangleq \sum_{i=1}^N \log{q^i_\phi (z^i| s^i)} - \lambda \log{q^i_\psi (z^i|s^{\neg i})}
\end{equation}

Importantly, $q^i_\phi$ can be viewed as representations for state factors~\citep{choi2021variational}, and thus we can specify a fixed distribution for $q^i_\phi$ to obtain an optimization objective similar to goal-conditioned RL. 
This conversion allows us to (optionally) manually specify the target values for (a subset of) state factors for more focused exploration, which we demonstrate in our experiments on Tiago robot.



\subsection{Safety-Aware Latent Action Learning}
\label{ss:safe}

Naively optimizing the objective in Eq.~\ref{eq:intri_rew} provides the robot with no notion of \textbf{safety}, which can result in irreversible damage when deployed on real hardware.
To address this issue, \methodname{} incorporates \textit{universal safety regularizer} in the form of a safety reward function $r_\mathit{safe}$ that discourages unsafe behaviors. 
In principle, $r_\mathit{safe}$ can take any form. 
In practice, for our real robot experiments, we found that the same safety reward function can be used universally across all environments and tasks.
Specifically, our safety reward $r_\mathit{safe}$ consists of the following components:
(1) Penalizing large absolute actions; 
(2) Penalizing large relative changes in action; 
(3) Penalizing collisions; 
(4) Penalizing excessive force on the robot. 
We provide the detailed formulation for $r_\mathit{safe}$ in App.~\ref{app:safe_r}.
The final objective for learning the latent action space combines task-agnostic exploration with these safety considerations, as shown below:
\begin{equation} \label{eq:final}
r_{\mathit{latent}} = r_{\mathit{dusdi}} + r_{\mathit{safe}}
\end{equation}

We directly optimize Eq.~\ref{eq:final} by running online RL in simulation.


\begin{algorithm*}[ht]
\caption{FLA-SAC for Real-World Downstream Task Learning\\
\smallskip
\textit{\hspace{2em}Note: Omitting standard SAC steps (e.g., target network creation and updates) for simplicity.}}
\label{alg:fla-sac}
Initialize task policy $\pi_{\mathit{task}}(z \mid o)$, factored Q-functions $\{Q^i\}_{i=1}^m$\;
Load pre-trained latent action decoder $\pi_\mathit{dec}$\; 
Initialize replay buffer $\mathcal{D}$ with NumStartSteps of random transitions \tcp*{Standard SAC warmstart} 
Dependency matrix $\mathcal{B}$ $\leftarrow$ infer($\mathcal{D}$) \tcp*{Sec.~\ref{ss:ar_dep}} 
\For{$k \leftarrow 1$ \KwTo $task\_learning\_steps$}{
    $z \gets \pi_{\mathit{task}}(z \mid o)$ \;
    $r_\text{sum} \gets [0]^m$ \;
    \For{$\text{t} \leftarrow 1$ \KwTo $steps\_per\_skill$}{
        $(o_\text{dec}, r = [r^i]_{i=1}^m, o') \gets \texttt{robot.real\_world\_step}(\pi_\mathit{dec}(a \mid o_\text{dec}, z))$\; 
        $r_\text{sum} = r_\text{sum} + r$ \;
    }
Store $(o, z, r_\text{sum}, o')$ into replay buffer $\mathcal{D}$\;
    \For{$j \leftarrow 1$ \KwTo $utd\_ratio$}{
        Sample mini-batch $\{(o, z, r, o')\}$ from $\mathcal{D}$ with \textbf{small batch size}\;
        Update $Q^i(o, \mathcal{B}_i \odot z)$ with $r^i$ \textbf{in parallel} for all $i = 1, \dots, m$\;
        Update $\pi_{\mathit{task}}(\hat{z} \mid o)$ with $Q = \sum_i Q^i$ using SAC loss, with $\hat{z}$ sampled via \textbf{Gumbel-Softmax} (Eq.~\ref{eq:gum}) for differentiability\;
    }
}
\KwRet{$\pi_\mathit{task}$}
\end{algorithm*}

\subsection{Accelerating Latent Action Space Learning through Q-Decomposition}
\label{ss:dec}


When using reinforcement learning (RL) to optimize the intrinsic reward function defined in Eq.~\ref{eq:final}, standard RL algorithms treat the reward function as a black box and learn a single value function from the mixture of intrinsic reward terms.
While this approach may be sufficient for environments with few state factors, doing so for complex environments with many state factors (large $N$) often leads to suboptimal solutions.
An important reason is that the mixture of $O(N)$ reward terms leads inevitably to high variance in the reward, making the value of the Q function oscillate. 
Furthermore, the sum of reward terms obscures information about each term's value, which hinders credit assignment.
This combination slows down or even prevents convergence.

\methodname{} overcomes this issue by leveraging the fact that the intrinsic reward function in Eq.~\ref{eq:intri_rew} is a linear sum over each disentangled component, and that the safety reward is also linearly added on top. Given any composite reward function $r_{\text{total}} = \sum_{i=1}^m r^i $, we can prove that $Q_{\text{total},\pi}(s, a) = \sum_{i=1}^m Q^{i}_{\pi} (s, a)$ using the linearity of expectations:\footnote{For SAC, the update will include an entropy bonus which can be treated as yet another reward term that is added linearly with the current reward function.}
\begin{proof}
\begin{align}
\label{eq:qdec}
Q_{\text{total},\pi}(s, a)  &= \mathbb{E}_{\pi} [\sum_{t=0}^{\infty} \gamma^t r_{\text{total}, t}] \nonumber \\ 
&= \mathbb{E}_{\pi} [\sum_{i=1}^m \sum_{t=0}^{\infty}  \gamma^t r^i_{t}] \nonumber \\ 
&= \sum_{i=1}^m \mathbb{E}_{\pi} [\sum_{t=0}^{\infty} \gamma^t  r^i_{t}] \nonumber \\
&= \sum_{i=1}^m Q^{i}_{\pi} (s, a) 
\end{align}
\end{proof}
In other words, we can factorize the Q function for Latent Action reward in Eq.~\ref{eq:final} into multiple sub-components, where the intrinsic reward term associated with each particular latent action dimension, $r^i \triangleq \log{q^i_\phi (z^i| s^i)} - \lambda \log{q^i_\psi (z^i|s^{\neg i})}$, is associated with a particular $Q^{i}_{\pi}$.
During policy learning, we sum all disentangled Q functions together to recover the global critic, $Q_{\pi}$. 
Compared to learning $Q_{\pi}$ directly from all the reward terms, learning disentangled Q functions significantly reduces reward variance, allowing $Q_{\pi}$ to converge faster and more stably, as we will show in the experiments.



\section{\methodname{} Step 2: Sample-Efficient Learning of Downstream Tasks in the Real World}
\label{ss:ds}

Given the latent action decoder $\pi_{\mathit{dec}}(a | o_{\mathit{dec}}, z)$ learned in simulation (Sec.~\ref{ss:la}), the second step of \methodname{} learns a high-level task policy $\pi_{\mathit{task}}: \mathcal{O} \rightarrow \bZ$ that maps robot observations (e.g. camera image) to latent action $z$. 
This downstream task learning phase happens \textbf{entirely in the real-world}, where taking actions is costly and therefore sample efficiency is of critical concern.
Towards sample-efficient real-world RL for challenging robotics tasks, we propose Factorized Latent-Action Soft Actor-Critic (FLA-SAC), a novel off-policy RL algorithm based on Soft Actor-Critic~\citep{haarnoja2018soft}.
In Sec.~\ref{ss:fpl}, we introduce how FLA-SAC utilizes masked Q-decomposition to achieve efficient learning when facing \emph{challenging downstream task with complicated objectives}. In Sec.~\ref{ss:ar_dep}, we discuss how the dependency matrix required for masked Q-decomposition can be automatically extracted from data, making FLA-SAC fully autonomous. In Sec.~\ref{ss:additional_algo}, we introduce additional algorithmic changes to further improve the sample-efficiency of FLA-SAC, and in Sec.~\ref{ss:discrete_action} we enable FLA-SAC to support large discrete action spaces via Gumbel-Softmax.
We show the pseudo-code for FLA-SAC in Algo.~\ref{alg:fla-sac}.

\begin{figure*}
    \centering
    \includegraphics[width=0.95\linewidth]{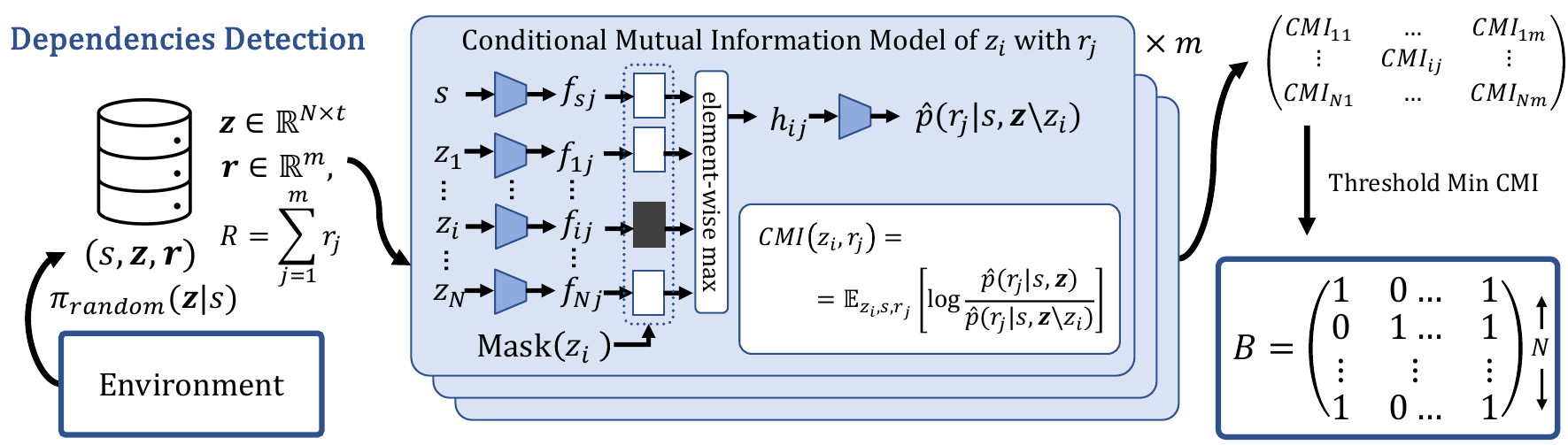}
    \caption{\methodname{} infers the dependencies between reward terms and latent action
dimensions through estimating and thresholding the conditional-mutual information (CMI) on randomly collected robot interaction data. To ensure computational efficiency, we introduce element-wise max as a mechanism to model different action conditioning.}
    \label{fig:discovery}
\end{figure*}

\subsection{Factorized Policy Learning via Masked Q-Decomposition}
\label{ss:fpl}

Challenging robotics tasks often come with a naturally \textit{composite} reward function, where the eventual reward is the sum of a set of reward terms corresponding to a set of sub-objectives, e.g. a whole-body mobile manipulation task may require: (1) navigating to a location, (2) without colliding with obstacles, and (3) while holding an object at the right orientation.
Optimizing this complex reward function with vanilla RL can be challenging, often requiring many steps of environment interactions, posing a significant challenge to real-world learning.
Instead, \methodname{} takes advantage of the \textbf{disentangled} nature of the learned latent action space $\bZ$, by \emph{finding and exploiting the strong dependencies between the latent action dimensions and the reward terms}.

Specifically, given a composite reward function with $m$ terms:
$r_\mathit{task} = \sum_{i=1}^m r^i$, we can first use the same Q-Decomposition in Eq.~\ref{eq:qdec} to derive that
   \textbf{ $Q_\pi(s, z) = \sum_{i=1}^m Q_\pi^i (s, z)$},
where each factored value function $Q_\pi^i$ represents the expected return for a specific reward term $r^i$.
Now, since each dimension $z^i$ of our latent action $z$ is trained to control and only control one environment entity, each reward term $r^i$ typically only depends on a \textit{small subset} of the latent action dimensions (for example, a reward for navigation is only associated with the latent action dimension that controls the robot base location). 
This property allows us to \textbf{dissociate each $Q_\pi^i$ with unrelated latent action dimensions} via masking, resulting in 
$Q_\pi^i(s, z^{\mathcal{I}_i})$, where  $\mathcal{I}_i \subseteq \{1, \dots, \dim(z)\}$ is the index set corresponding to the dimensions of $z$ that reward $r^i$ depends on. 

Formally, we represent these action-reward dependencies as a binary adjacency matrix $\mathcal{B} \in \{0,1\}^{m \times N}$, where $B_{ij}=0$ if and only if 
\begin{equation}
    R^i
    \;\perp\!\!\!\perp\;
    {\bZ}^j
    \;\Big|\;
    s_0,\; \bZ / {\bZ}^j
    \label{eq:disentangle_assump}
\end{equation}
for any $s_0$. In other words, the $i$-th reward term is conditionally independent of (and therefore not affected by)
latent action dimension $j$, given the state and the rest of the latent action dimensions. 
Therefore, we can use this dependency matrix $\mathcal{B}$ to mask out irrelevant latent action dimensions for each $Q_\pi^i$.

Masked Q-Decomposition brings us two significant advantages: First, it effectively prevents $Q_\pi^i$ from learning spurious correlations with latent actions, thereby providing a better optimization landscape for accurate prediction of the Q value. Second, since we calculate the policy update by backpropagating from the Q functions, each action dimension will only be updated with reward terms that it can affect, resulting in a more accurate policy gradient estimation than the non-factorized version. In other words, our technique effectively decomposes a hard learning problem into multiple simpler problems that can be solved in parallel, leading to improved performance and sample efficiency.

While the matrix $\mathcal{B}$ can often be manually specified without much effort, we introduce a pipeline for automatically learning this dependency matrix from data in Sec.~\ref{ss:ar_dep}, which allow us to fully automate the downstream learning.



\subsection{Autonomous Detection of Action-Reward Dependencies}
\label{ss:ar_dep}

In this section, we introduce a method for autonomously inferring action-reward dependencies from robot interaction data (Fig.~\ref{fig:discovery}), inspired by our previous work~\citep{Hu-RSS-23}. Importantly, \textbf{this inference step does not require extra samples}, since the interaction data is also used for initializing the replay buffer of the RL agent.


Given this random interaction dataset, we infer the dependency matrix $\mathcal{B}$ through conditional independence tests, which can be made by measuring the Conditional Mutual Information (CMI) between pairs of action space dimensions and reward terms, $(i,j)$, as follows:
\begin{equation}
\begin{aligned} 
    \text{CMI}(z_i,r_j) &= \mathbb{E}_{z_i, s, r_j} \left[log\frac{p(z_i, r_j|\{s, \mathbf{z} \backslash z_i\})}{p(z_i|\{s, \mathbf{z} \backslash z_i\}) p(r_j|\{s, \mathbf{z} \backslash z_i\})}\right] \\&=\mathbb{E}_{z_i, s, r_j} \left[ log \frac{p(r_j | \{s, \mathbf{z} \})}{p(r_j | \{s, \mathbf{z} \backslash z_i\})}\right]
\end{aligned}
\end{equation}
where the expectation is taken over the joint distribution of $\{z_i, s, r_j\}$. We consider that a dependency edge $z_i \rightarrow r_j$ exists if $\text{CMI}(z_i,r_j) > \epsilon$, where $\epsilon$ is a mutual-information threshold. 

We estimate the CMI between action dimensions and reward terms by training predictive models, $\hat{p}(r_j | \{s, \mathbf{z} \})$ and $\hat{p}(r_j | \{s, \mathbf{z} \backslash z_i\}$, for each $i, j$ over the exploratory dataset. However, training a separate model for estimating each probability would require a total of $N \times m$ models, which is computationally costly. To efficiently estimate the CMI between actions and rewards, we adopt the model architecture and training procedure proposed by \citet{wang2022causal}, originally used for learning causal dynamic models, in the form explained below. This model architecture reduces the total number of models needed to $m$, the number of reward terms.

Specifically, for each reward channel $r_j$, we train a model (Fig.~\ref{fig:discovery}) that predicts the value of that reward channel from a full or partial action vector and the state, $\hat{p}(r_j | \{s, \mathbf{z} \})$ and $\hat{p}(r_j | \{s, \mathbf{z} \backslash z_i\}$. The model consists of three steps: first, each of the action dimensions, $z_1, ..., z_n$, and the state $s$ are individually mapped to feature vectors $f_{1,j}(z_1), ..., f_{n,j}(z_n), f_{s,j}(s)$ of equal length $l$ ($l=128$ in this work). Then, an overall feature $h_j$ is obtained by taking the element-wise max of all features. A prediction network $g_j()$ maps $h_j$ to the predicted reward channel. Using all values of the action vector as input, this procedure approximates the full conditional probability, $g_j(h_j)=\hat{p}(r_j | \{s, \mathbf{z} \})$. To estimate the conditional probability of the conditioning set for the action dimension $z_i$, we use a mask that sets the feature corresponding to $z_i$, $f_{i,j}(z_i)$, to $-\infty$ and repeat the reward inference obtaining $g_j(h_j)=\hat{p}(r_j | \{s, \mathbf{z} \backslash z_i\}$.
This model is trained to maximize the following log-likelihood:
\begin{equation}
\label{eqn:train_obj}
\begin{aligned} 
L = \sum_j [\log \hat{p}( r_j|s, \mathbf{z}) &+ \log \hat{p} (r_j | \{s, \mathbf{z} \backslash z_i\})]
\end{aligned}
\end{equation}
where $i$ is uniformly sampled from $\{1, \ldots, n \}$ for each $j$.
Maximizing equation~\ref{eqn:train_obj} corresponds to maximizing the accuracy of the two terms necessary for estimating $\text{CMI}(z_i,r_j)$, which promotes an accurate estimation of the dependency graph. 

We split the interaction data into the training part for maximizing $L$ and the validation part for evaluating CMI.
After training, we obtain the bi-adjacency dependency matrix ${B}$ by examining the CMI for each reward-action pair, $(i,j)$, based on the model's predicted conditional probability for each reward term in the validation dataset. 


\begin{figure*}[t]
    \centering
    \begin{subfigure}[t]{0.25\textwidth}
        \hspace*{0.16\textwidth}
        \includegraphics[width=0.8\textwidth]{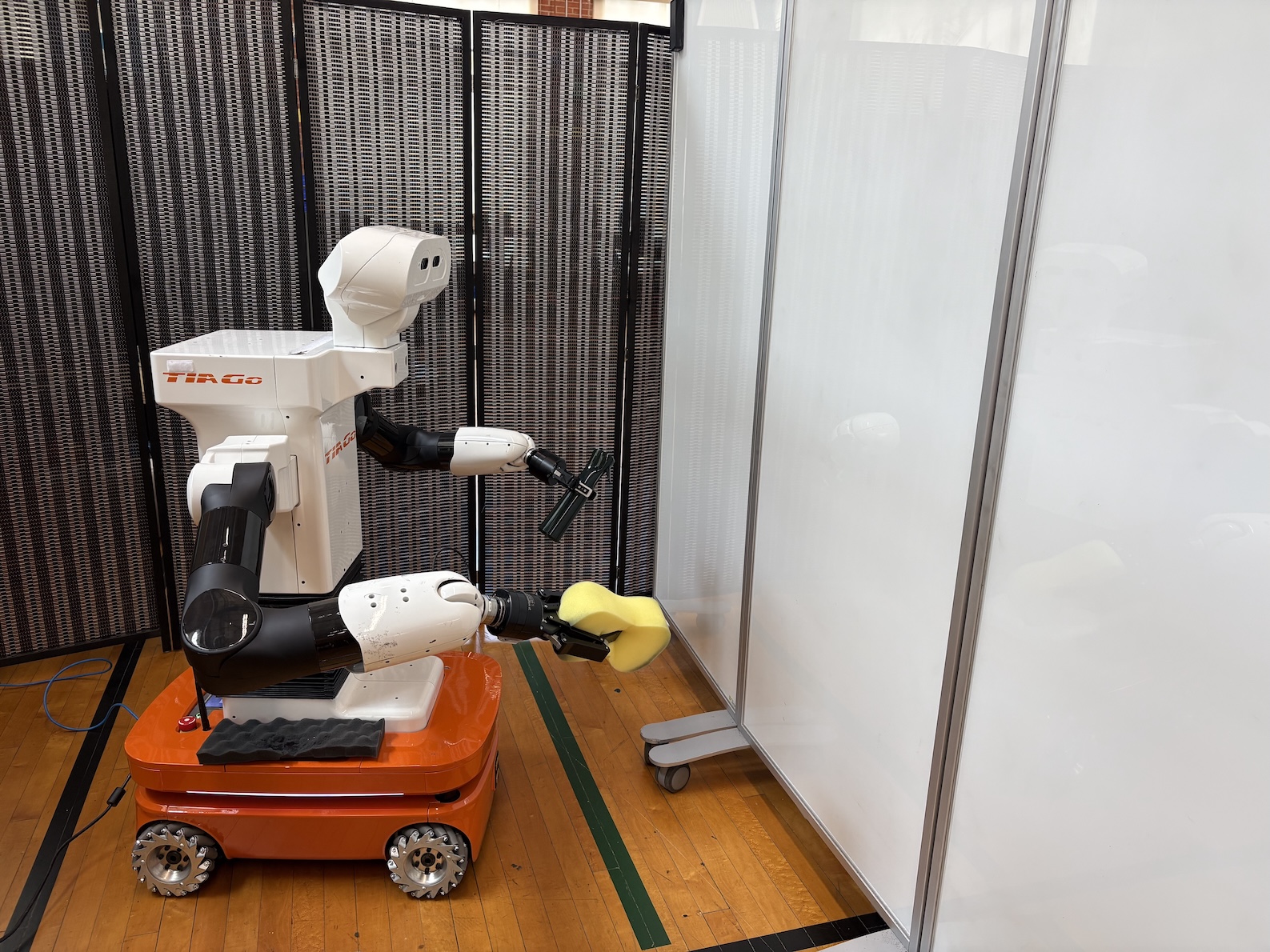}
        \vspace{0.5em}
    \end{subfigure}%
    \begin{subfigure}[t]{0.25\textwidth}
        \hspace*{0.16\textwidth}
        \includegraphics[width=0.8\textwidth, trim=0 55 0 40, clip]{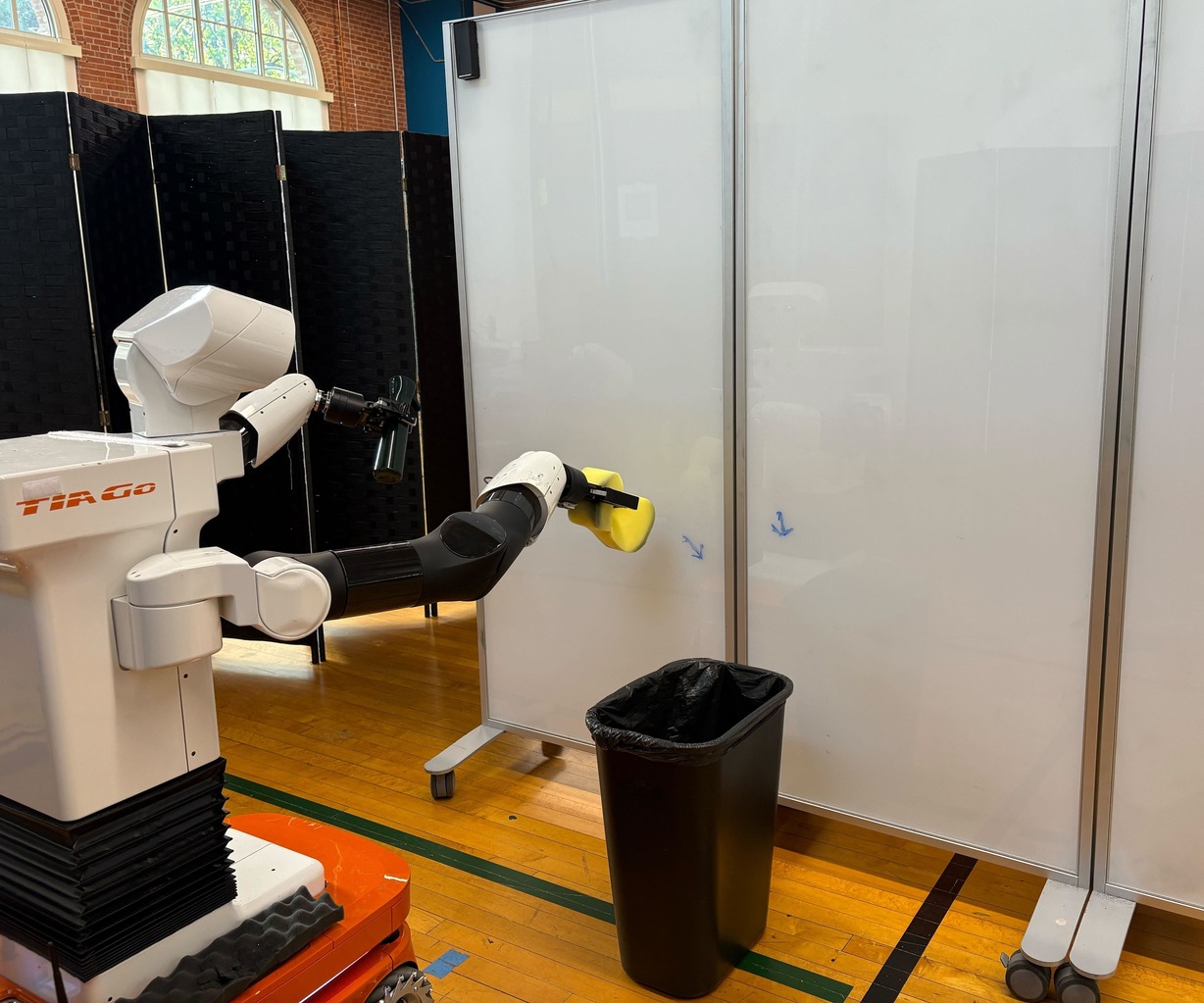}
        \vspace{0.5em}
    \end{subfigure}%
    \begin{subfigure}[t]{0.25\textwidth}
        \hspace*{0.16\textwidth}
        \includegraphics[width=0.8\textwidth, trim=0 15 0 5, clip]{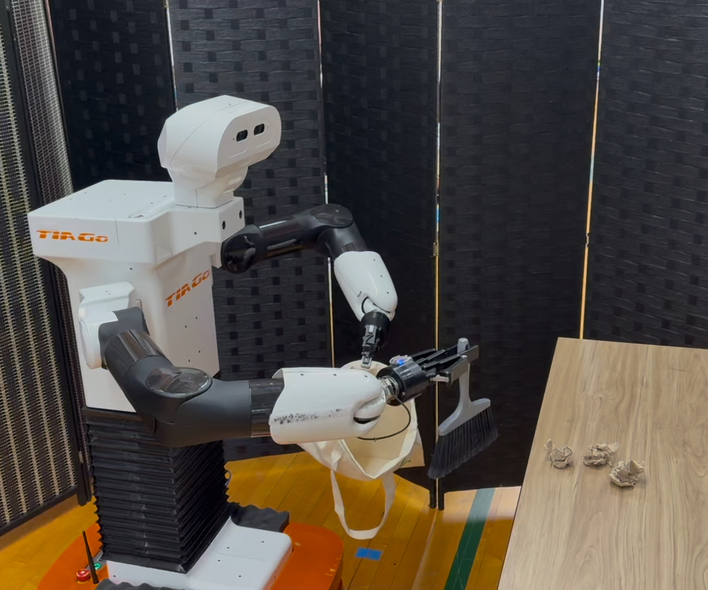}
        \vspace{0.5em}
    \end{subfigure}%
    \begin{subfigure}[t]{0.25\textwidth}
        \hspace*{0.16\textwidth}
        \includegraphics[width=0.8\textwidth]{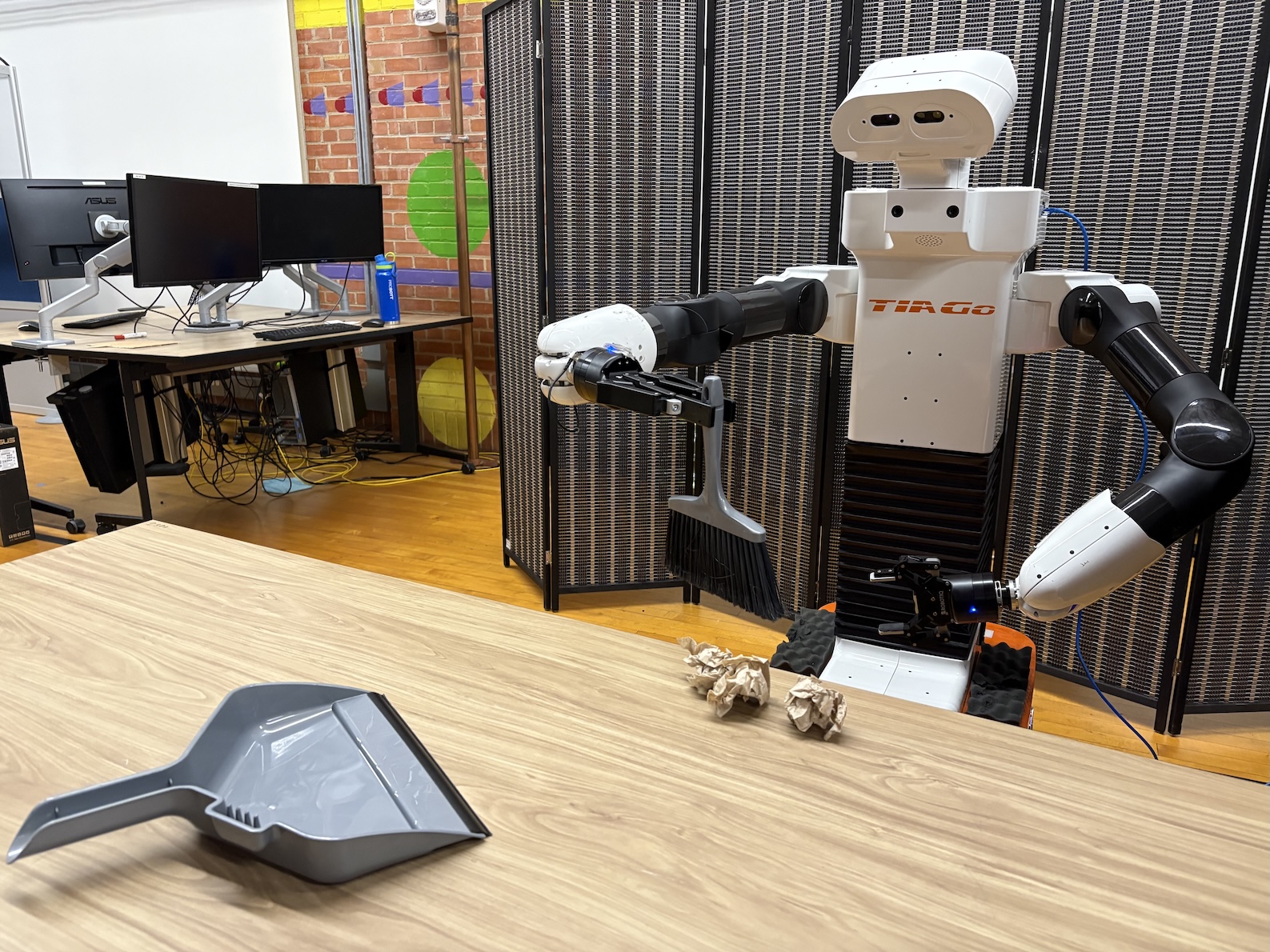}
        \vspace{0.5em}
    \end{subfigure}
    
    \begin{subfigure}[t]{0.24\textwidth}
        \centering
        \includegraphics[width=\textwidth, trim=15 25 15 20, clip]{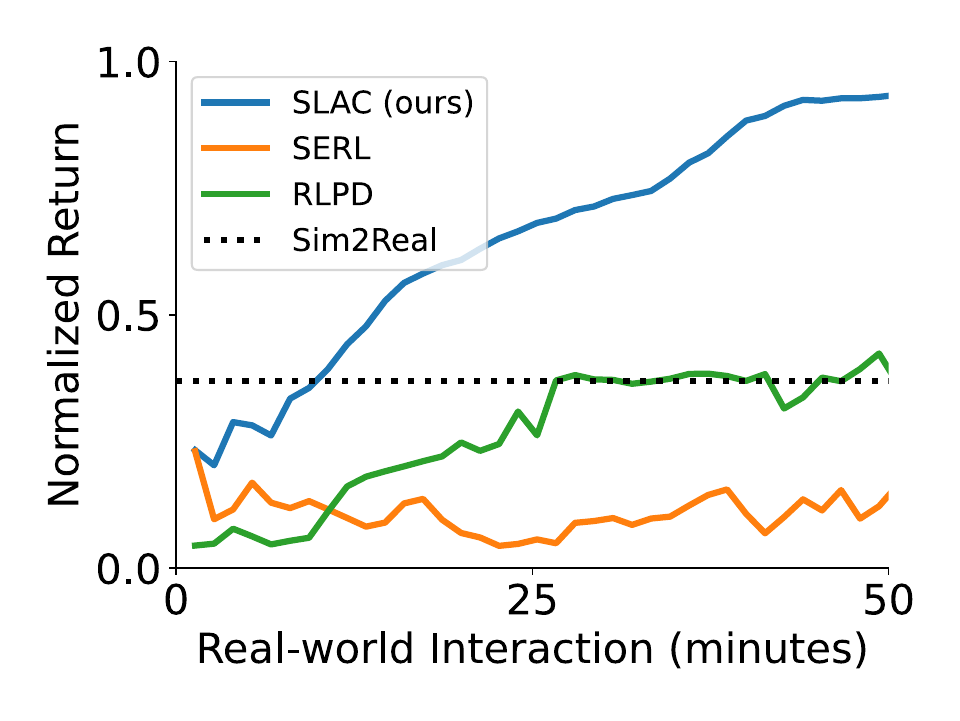}
        \caption{Wipe Board Task}
    \end{subfigure}
    \begin{subfigure}[t]{0.24\textwidth}
        \centering
        \includegraphics[width=\textwidth, trim=15 25 15 20, clip]{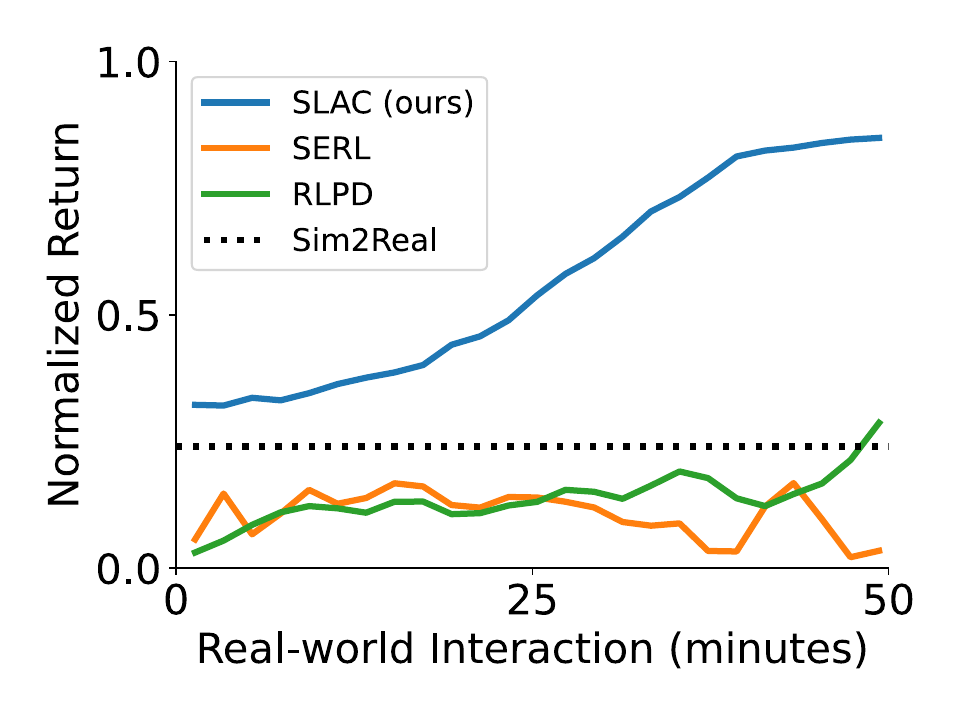}
        \caption{Board with Obstacle Task}
    \end{subfigure}
    \begin{subfigure}[t]{0.24\textwidth}
        \centering
        \includegraphics[width=\textwidth, trim=15 25 15 20, clip]{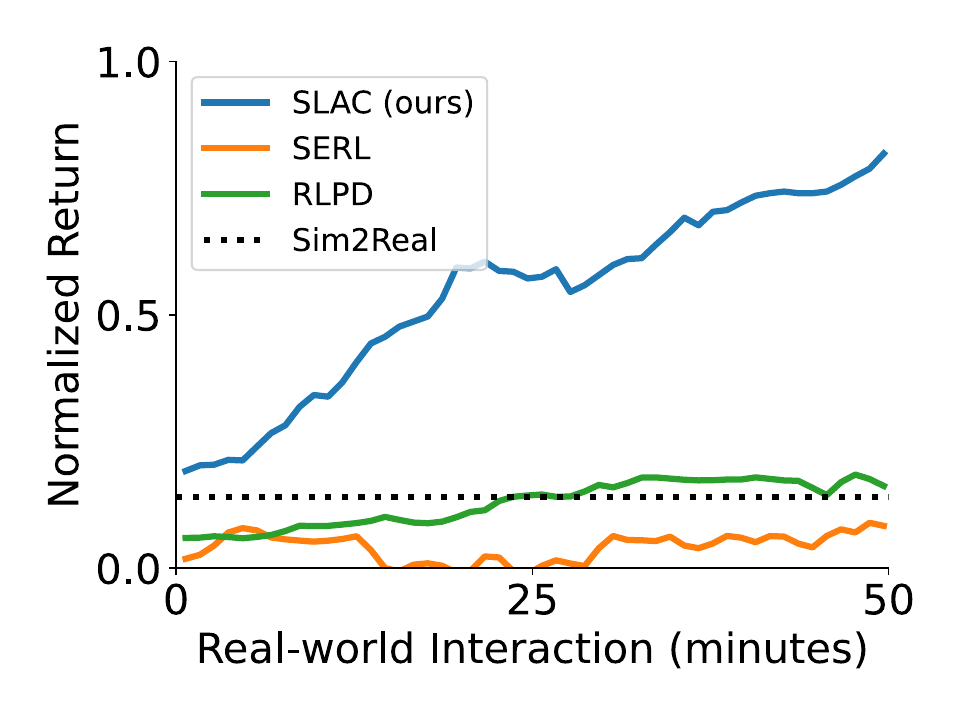}
        \caption{Trash to Bag Task}
    \end{subfigure}
    \begin{subfigure}[t]{0.24\textwidth}
        \centering
        \includegraphics[width=\textwidth, trim=15 25 15 20, clip]{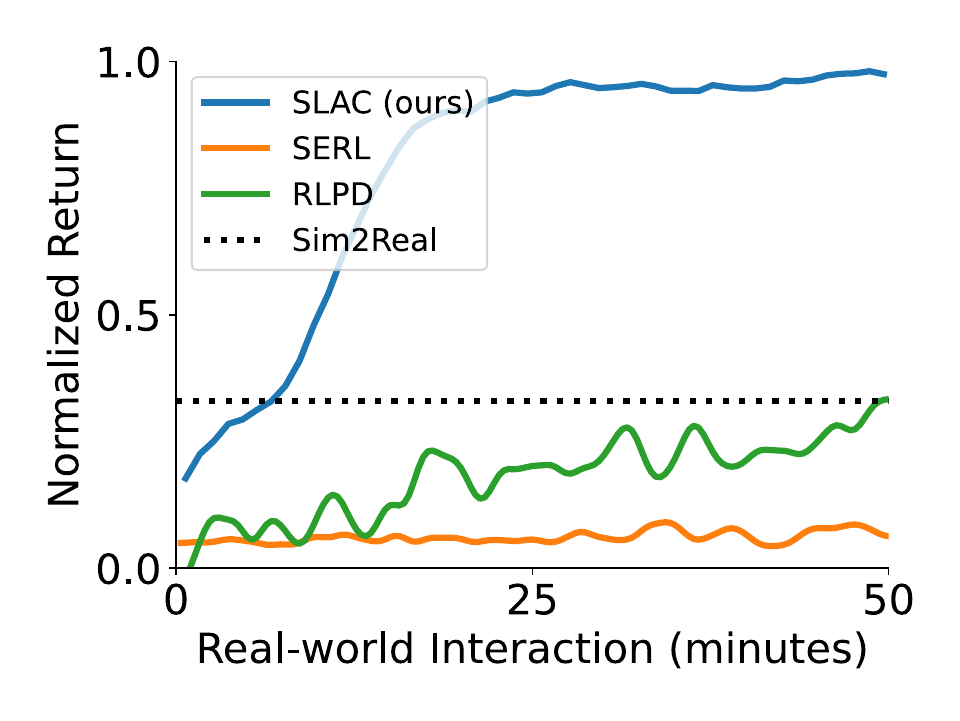}
        \caption{Trash to Tray Task}
    \end{subfigure}

    \caption{ Training curves for \methodname{} and baseline methods. \methodname{} can learn challenging contact-rich whole-body mobile manipulation tasks within an hour of real-world interactions, significantly outperforming prior methods.
    }
    \label{fig:rslt}
\end{figure*}

\begin{table*}[t]
\centering
\renewcommand{\arraystretch}{1.2}
\setlength{\tabcolsep}{6pt}
\caption{We compare the number of successful trials ($\uparrow$) over 10 rollouts \textbf{after training} and the total safety violation counts ($\downarrow$) \textbf{during training} between \methodname{} and baseline methods across four tasks. In all four tasks, \methodname{} achieves the highest success rate while also inducing the least number of safety violations.}
\footnotesize
\begin{tabular}{lcccccccc}
\toprule
\textbf{Method /\ Task} 
& \multicolumn{2}{c}{\textbf{Board}} 
& \multicolumn{2}{c}{\textbf{Board-Obstacle}} 
& \multicolumn{2}{c}{\textbf{Table-Tray}} 
& \multicolumn{2}{c}{\textbf{Table-Bag}} \\
\cmidrule(lr){2-3} \cmidrule(lr){4-5} \cmidrule(lr){6-7} \cmidrule(lr){8-9}
& Success \# & Unsafe \# 
& Success \# & Unsafe \# 
& Success \# & Unsafe \# 
& Success \# & Unsafe \# \\
\midrule
\methodname{} (ours) & \textbf{9 / 10} & \textbf{1} & \textbf{8 / 10} & \textbf{4} & \textbf{9 / 10} & \textbf{0} & \textbf{7 / 10} & \textbf{0} \\
SERL~\citep{luo2024serl} & 0 / 10 & 8 & 0 / 10 & 22 & 0 / 10 & 6 & 0 / 10 & 9 \\
Sim2Real~\citep{tobin2017domain} & 2 / 10 & - & 2 / 10 & - & 4 / 10 & - & 0 / 10 & - \\
RLPD~\citep{ball2023efficient} & 4 / 10 & 34 & 2 / 10 & 37 & 3 / 10 & 26 & 0 / 10 & 33 \\
\bottomrule
\end{tabular}
\label{tab:success_rates}
\end{table*}


\subsection{Further Boosting Sample Efficiency via High UTD Ratio and Regularization}
\label{ss:additional_algo}
Due to the high cost of collecting real-world trajectories, our goal is to develop algorithms that efficiently learn from a few steps of environment interactions. Aside from the masked Q-decomposition, one critical strategy to achieve this efficiency is giving an off-policy algorithm a high update-to-data (UTD) ratio, where the number of actor-critic updates is significantly higher than the number of environment steps, by repeatedly sampling from a replay buffer that stores all previous environment steps. FLA-SAC leverages such a high UTD ratio to maximize data efficiency. 

Since a high UTD ratio can increase the risk of overfitting, recent work~\citep{smith2022walk,ball2023efficient,luo2024serl} proposed various techniques, such as layer normalization and critic ensembling, to regularize the policy update. However, we empirically found these methods to be ineffective in our setting.
Instead, we observed that simply reducing the batch size during updates (from 256 in the original SAC implementation~\citep{haarnoja2018soft} to 64) acts as powerful regularization that significantly improves performance.
This adjustment helps the model escape poor local optima by introducing higher gradient variance, which promotes more effective exploration of the parameter space, especially at the start of the training.


\subsection{Handling Large Discrete Action Space}
\label{ss:discrete_action}
Since we opt for a discrete latent action space in \methodname{}, we want our downstream learning algorithm to support discrete actions. Unfortunately, vanilla SAC only works for continuous actions due to the need to backpropagate through the action vector during policy update. While there exist off-policy algorithms that support discrete action spaces (e.g.  DQN~\citep{mnih2015human}, Discrete-SAC~\citep{christodoulou2019soft}), they typically require enumerating the Q function for all possible actions and do not work for combinatorially large discrete action spaces (e.g. the latent action space of \methodname{}). 

Instead, FLA-SAC extends SAC to large discrete action spaces by using the gumbel-softmax trick~\citep{jang2016categorical}, which allow us to compute gradients through discrete random variables via reparametrization. Specifically, we used the non-straight-through Gumbel-softmax estimation shown in Eq.~\ref{eq:gum} for sampling actions, with a fixed temperature $\tau$ of 1.0, which we empirically found to give good performance even when the size of the action space is as large as $\mathcal{O}(10^6)$.

\begin{equation}
\begin{aligned}
    \hat{z}(s) = &\mathit{softmax}\left(\frac{\log \pi_\theta(z \mid s) + g_z}{\tau}\right), \\
   & g_z \sim \mathit{Gumbel}(0, 1)
\end{aligned}
\label{eq:gum}
\end{equation}

\section{Experimental Results}

In this section, we conduct a series of experiments to examine the performance \methodname{}. In Sec.~\ref{exp:moma}, we show how \methodname{} can enable a Tiago mobile manipulator to learning contact-rich visuo-motor tasks through real-world interactions. In Sec.~\ref{ss:par}, we illustrate the general applicability of the \methodname{} framework by testing it on a multi-robot domain. Finally, in Sec.~\ref{ss:abla}, we perform ablation studies to understand the impact of each component on the final performance.



\subsection{\methodname{} for Real-World Mobile Manipulation}
\label{exp:moma}
Our primary experiments evaluate \methodname{} on a bi-manual mobile manipulator -- a domain that is ideally suited to our approach. This domain is not only highly challenging due to the complexity of the embodiment and task space, but also practically important for the development of capable household robots.
Specifically, we evaluate \methodname{} in two different environments: a table environment, where the robot faces a table with objects on it; and a whiteboard environment, where the robot can interact with a whiteboard. Each simulation environment takes less than 20 minutes to create in iGibson \citep{li2022igibson}, using off-the-shelf object models without any real2sim.
In each environment (shown in Fig.~\ref{fig:env_viz}, described in detail in Appendix~\ref{app:env_moma}), we evaluate multiple different \textbf{visuomotor contact-rich} tasks that require \textbf{whole-body} motion to solve.

\textbf{Observations \& Network}: In all tasks, the observation of the robot consists \textbf{only} of an RGBD camera observation and robot proprioception. The pointcloud is first processed by a PointNet~\citep{qi2017pointnetdeeplearningpoint} and then passed into an MLP network along with the proprioceptive data. Both networks are randomly initialized and trained from scratch. We show the detailed hyperparameters in the appendix.





\begin{figure*}[t]
\centering

\begin{subfigure}{0.24\textwidth}
    \includegraphics[width=\linewidth]{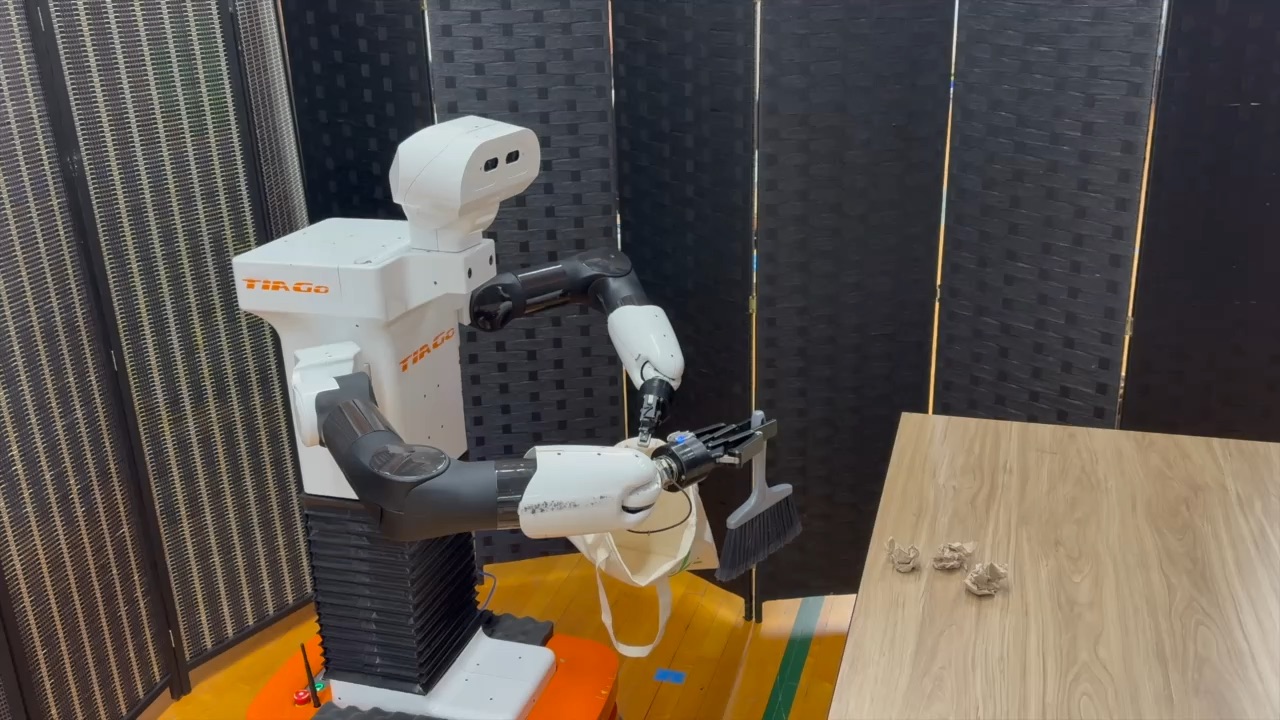}
\end{subfigure}
\hfill
\begin{subfigure}{0.24\textwidth}
    \includegraphics[width=\linewidth]{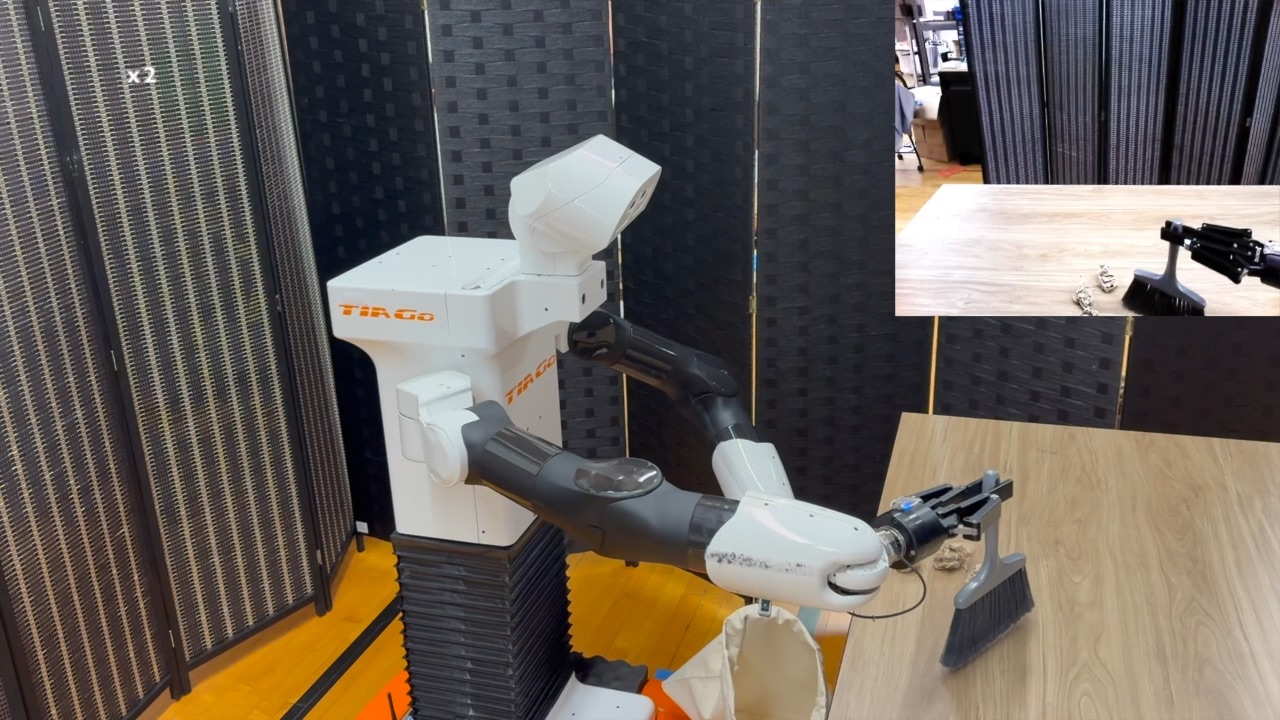}
\end{subfigure}
\hfill
\begin{subfigure}{0.24\textwidth}
    \includegraphics[width=\linewidth]{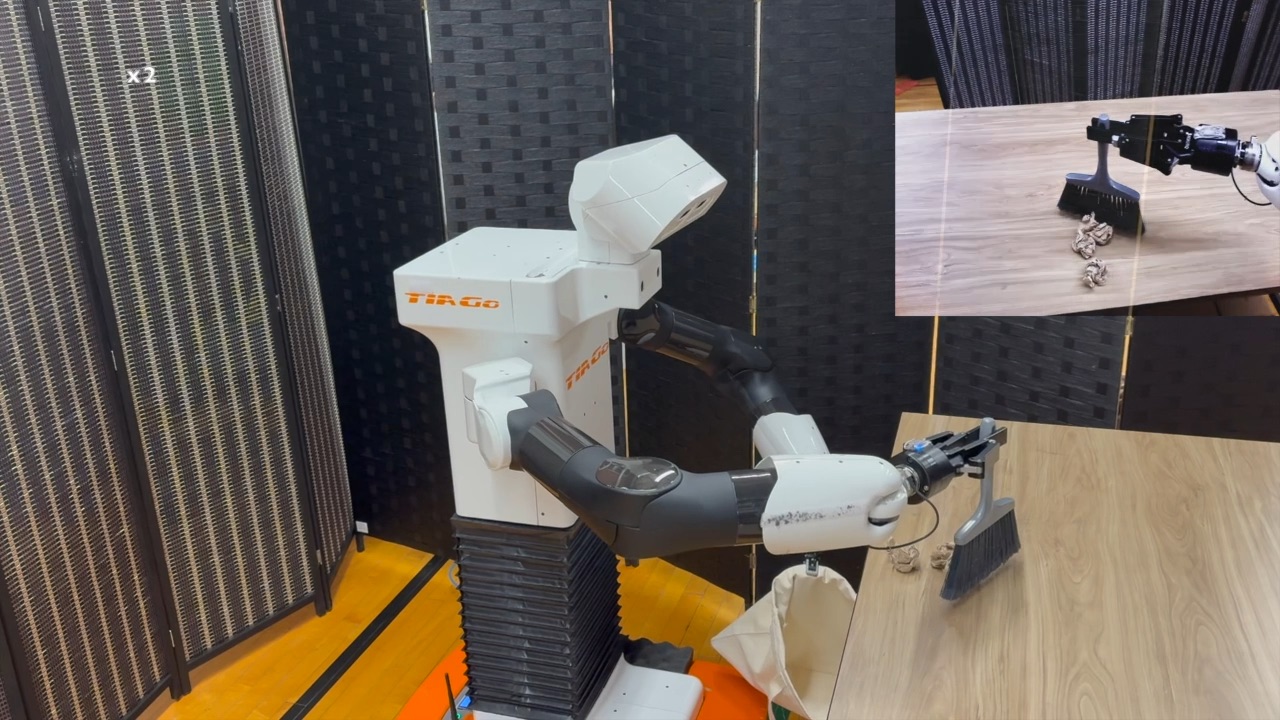}
\end{subfigure}
\hfill
\begin{subfigure}{0.24\textwidth}
    \includegraphics[width=\linewidth]{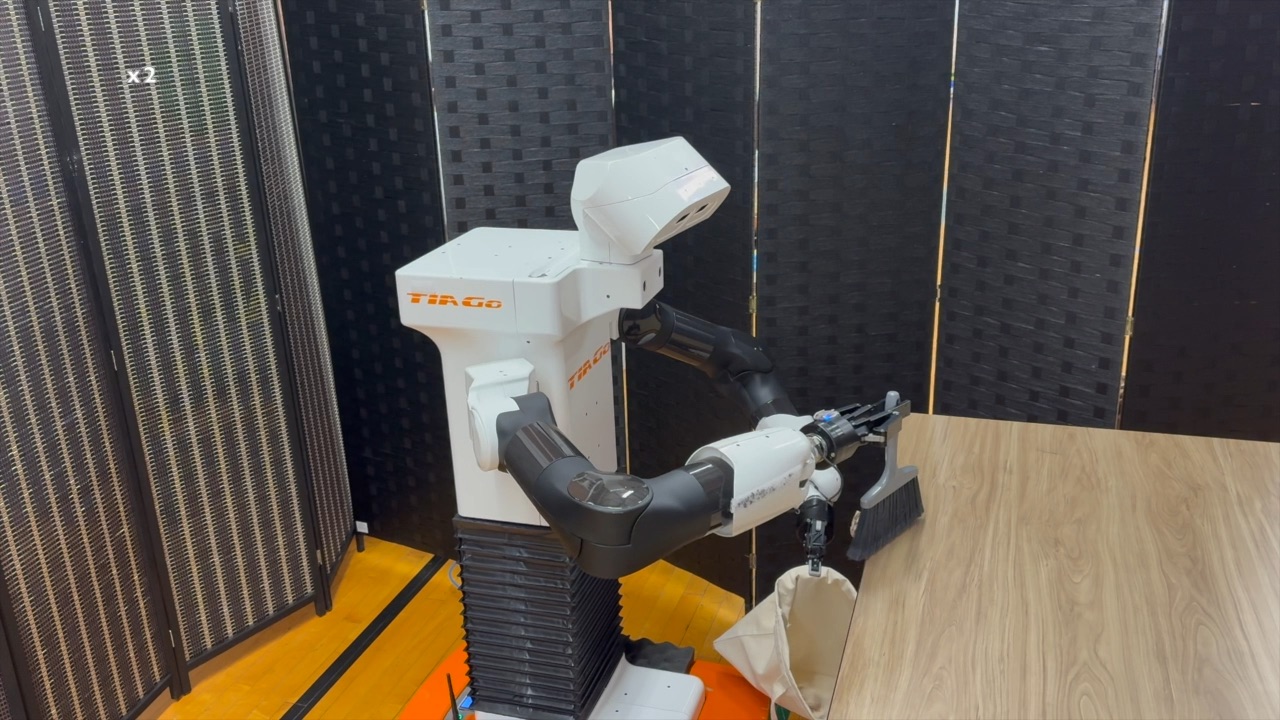}
\end{subfigure}
\caption{Motion sequence of the ``Trash to Bag'' task (left to right). The robot approaches the table and sweeps the trash into the bag that it is holding.}
\label{fig:motion_sequence_bag}
\end{figure*}

\begin{figure*}[t]
\centering

\begin{subfigure}{0.24\textwidth}
    \includegraphics[width=\linewidth]{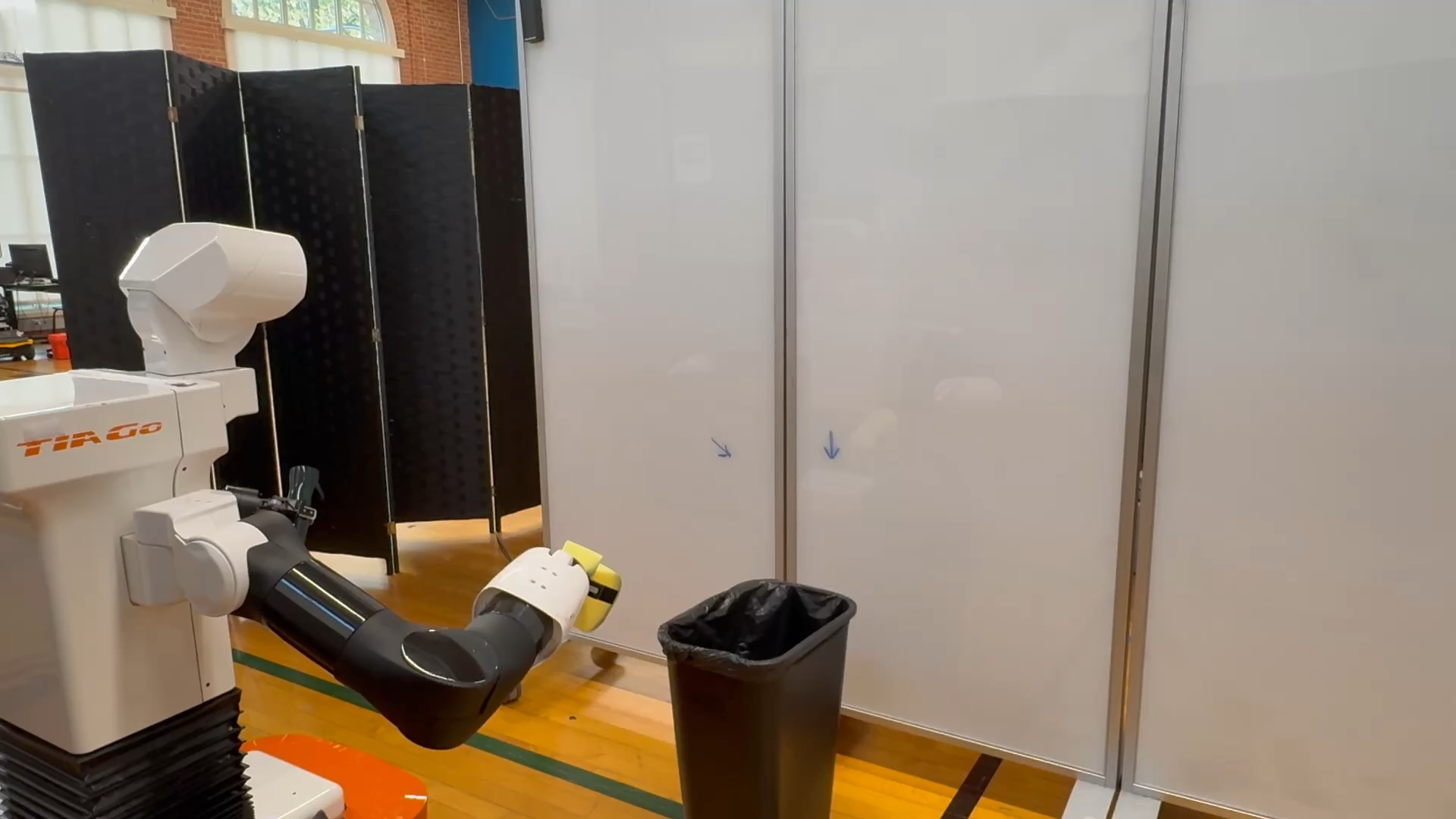}
\end{subfigure}
\hfill
\begin{subfigure}{0.24\textwidth}
    \includegraphics[width=\linewidth]{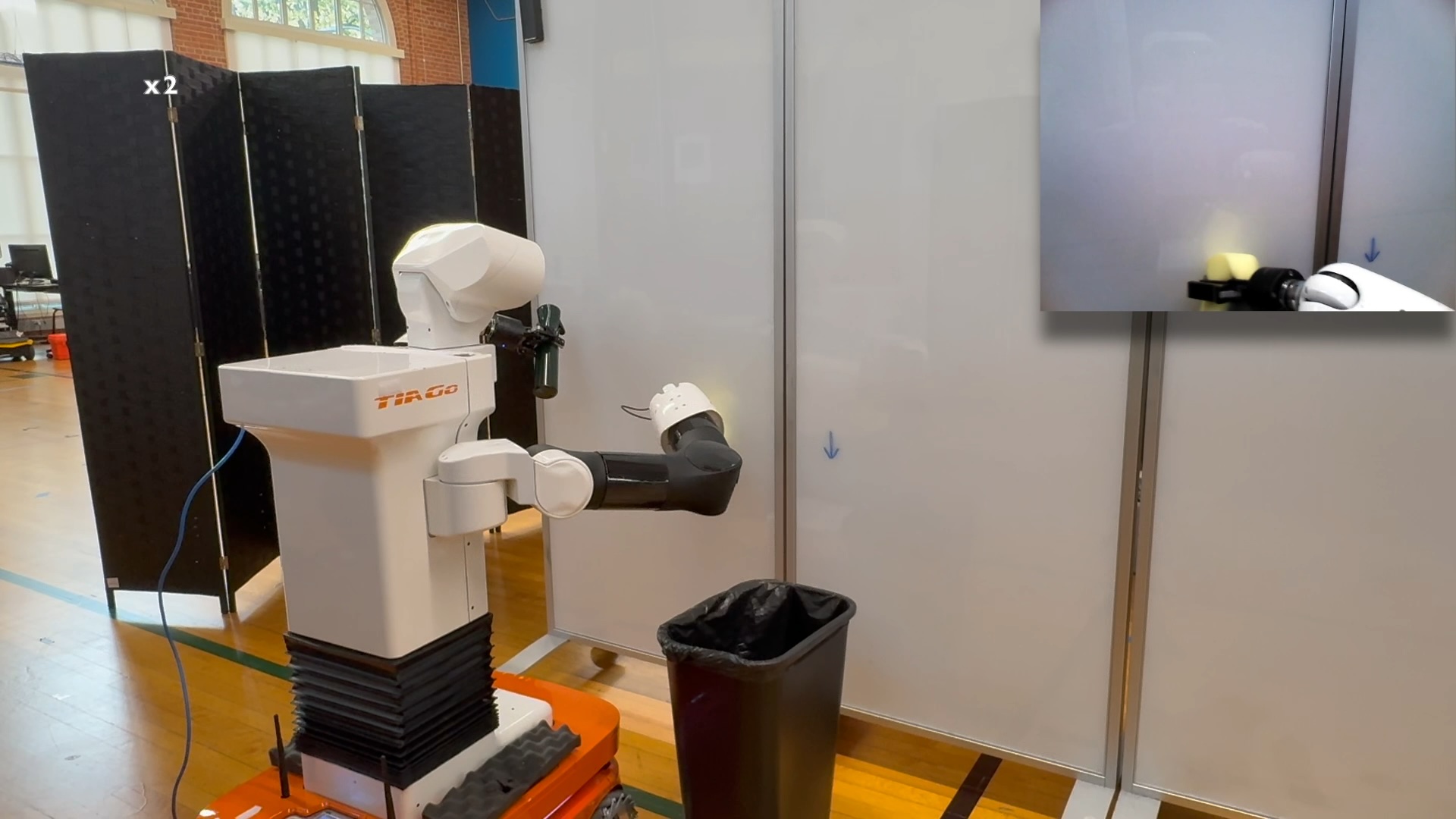}
\end{subfigure}
\hfill
\begin{subfigure}{0.24\textwidth}
    \includegraphics[width=\linewidth]{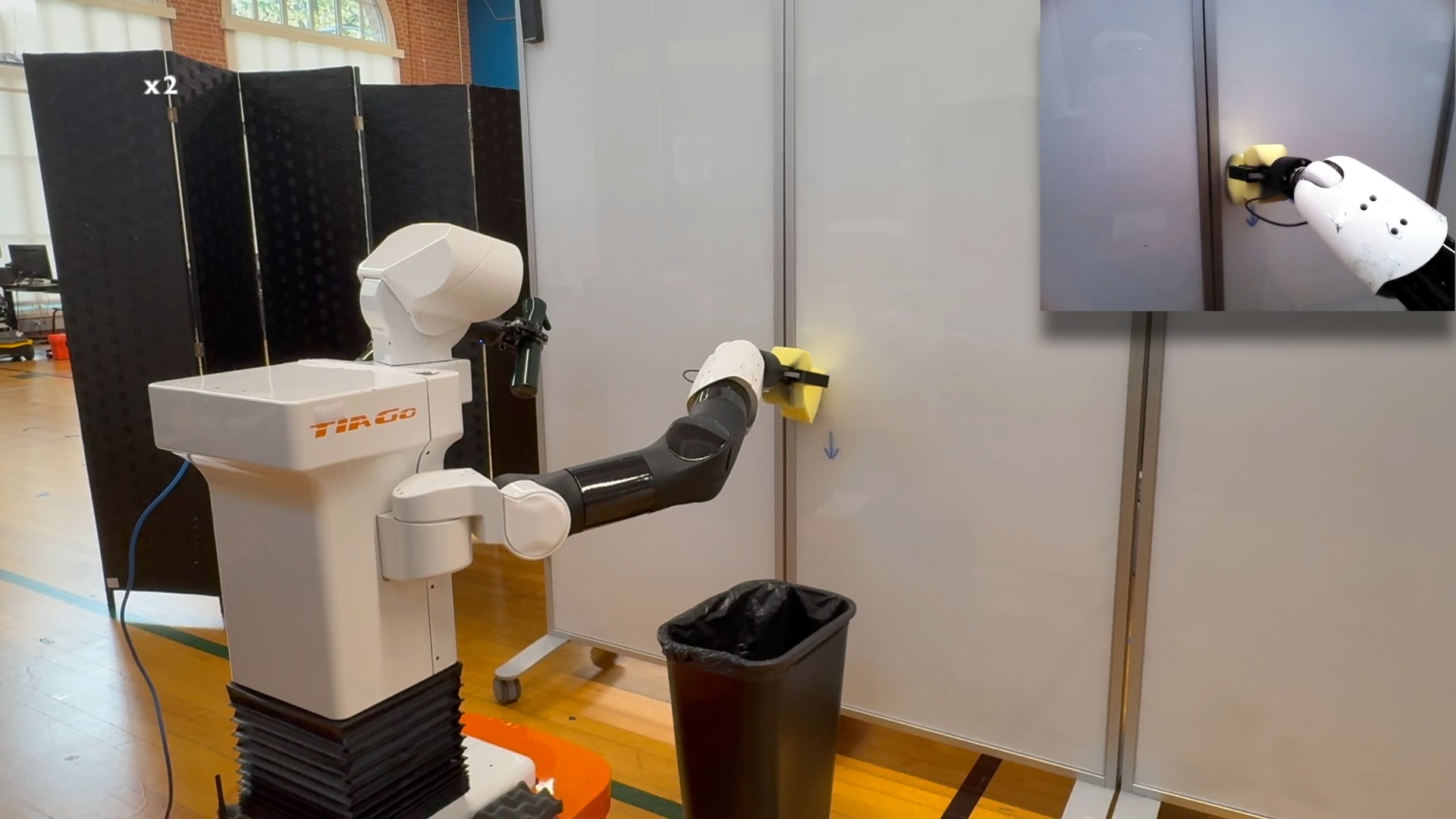}
\end{subfigure}
\hfill
\begin{subfigure}{0.24\textwidth}
    \includegraphics[width=\linewidth]{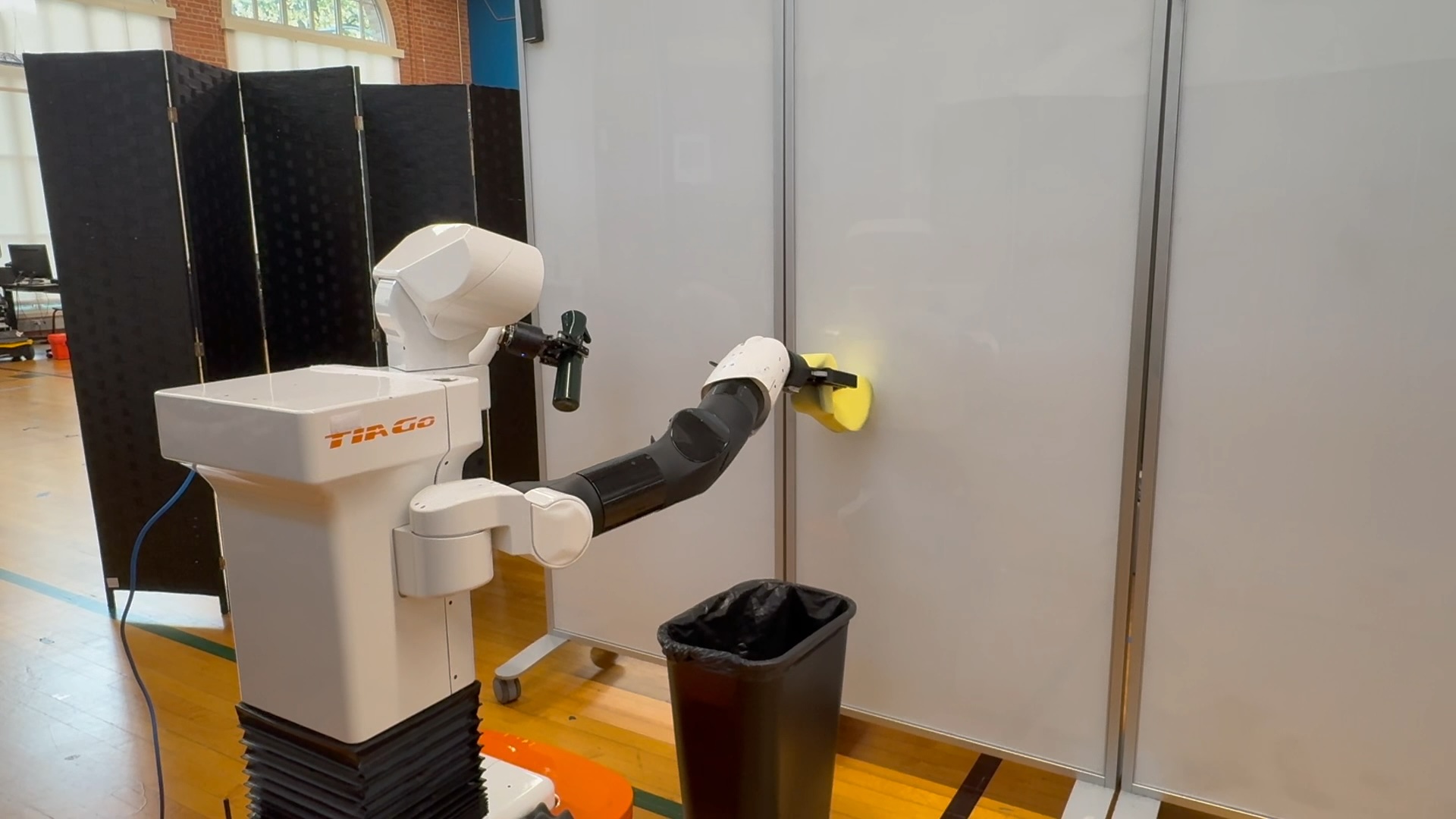}
\end{subfigure}
\caption{Motion sequence of the ``Board with Obstacle'' task (left to right). The robot wipes the mark on the whiteboard while positioning itself to avoid colliding with the obstacle.}
\label{fig:motion_sequence_board}
\end{figure*}

\textbf{Baselines:}
We compare the performance of \methodname{} against state-of-the-art methods in real-world RL, sim-to-real RL, and real-world finetuning of simulation policy. Specifically, we compare against:

\begin{itemize}[topsep=-1pt,leftmargin=*]
\setlength{\itemsep}{0pt}
\setlength{\parskip}{0pt}
\setlength{\parsep}{0pt}
    \item \textbf{SERL} \emph{(Real-World RL from Scratch)}~\citep{luo2024serl}, a state-of-the-art real-world RL framework that directly train a policy from scratch in the low-level action space using regularized SAC.
    \item \textbf{Zero-shot Sim2Real}~\citep{tobin2017domain}, a task policy trained in sim is directly applied to the real world, with domain randomization applied to the robot observation. 
    \item \textbf{RLPD} \emph{(Sim2Real with Real-World Finetuning)}~\citep{ball2023efficient}, a state-of-the-art method for learning from both online and prior data, which we use for finetuning in the real world with prior data from simulation.
\end{itemize}

Notice that \emph{both Zero-shot Sim2Real and RLPD have an unfair advantage} over \methodname{}, as they require implementing downstream task reward and objects (e.g. marker trace) in simulation. For contact-rich tasks, these objects are often quite hard to create in simulation. 
By comparison, \methodname{} does not require implementing the exact downstream tasks in simulation, since we are only learning a task-agnostic latent action space that can facilitate safe exploration.
Furthermore, these methods require separate training for each task, while \methodname{} can reuse the same latent action space for different tasks in the same scene.

\textbf{Evaluation Metric}: For each method on each downstream task, we compare the success rate of the final policy across 10 rollouts with different initial states. For the three methods that require training in simulation (\methodname{}, Sim2Real, RLPD), we train for 10M steps in simulation. For the three methods that require real-world interactions (\methodname{}, SERL, RLPD), we train each of them for 30k steps of real-world low-level robot actions, corresponding to 50 minutes of real-world interactions, and additionally report the number of times they have violated the safety constraints during training. 

\textbf{Results}: The full results are shown in Table~\ref{tab:success_rates}. In all four tasks, \methodname{} learns to solve the task in less than an hour of real-world interactions (curves in Fig.~\ref{fig:rslt}), while maintaining safety during real-world exploration, significantly outperforming the baselines. 
These results showcase the capabilities of \methodname{} for learning complicated tasks in domains that are extremely challenging for previous Reinforcement Learning methods.
Qualitatively, we show some of the robot trajectories in Fig.~\ref{fig:motion_sequence_bag} and Fig.~\ref{fig:motion_sequence_board}.

\subsection{\methodname{} for Multi-Robot Systems}
\label{ss:par}

\begin{figure}[h]
    \centering

    \begin{subfigure}[t]{0.13\textwidth}
        \centering
        \includegraphics[width=\textwidth]{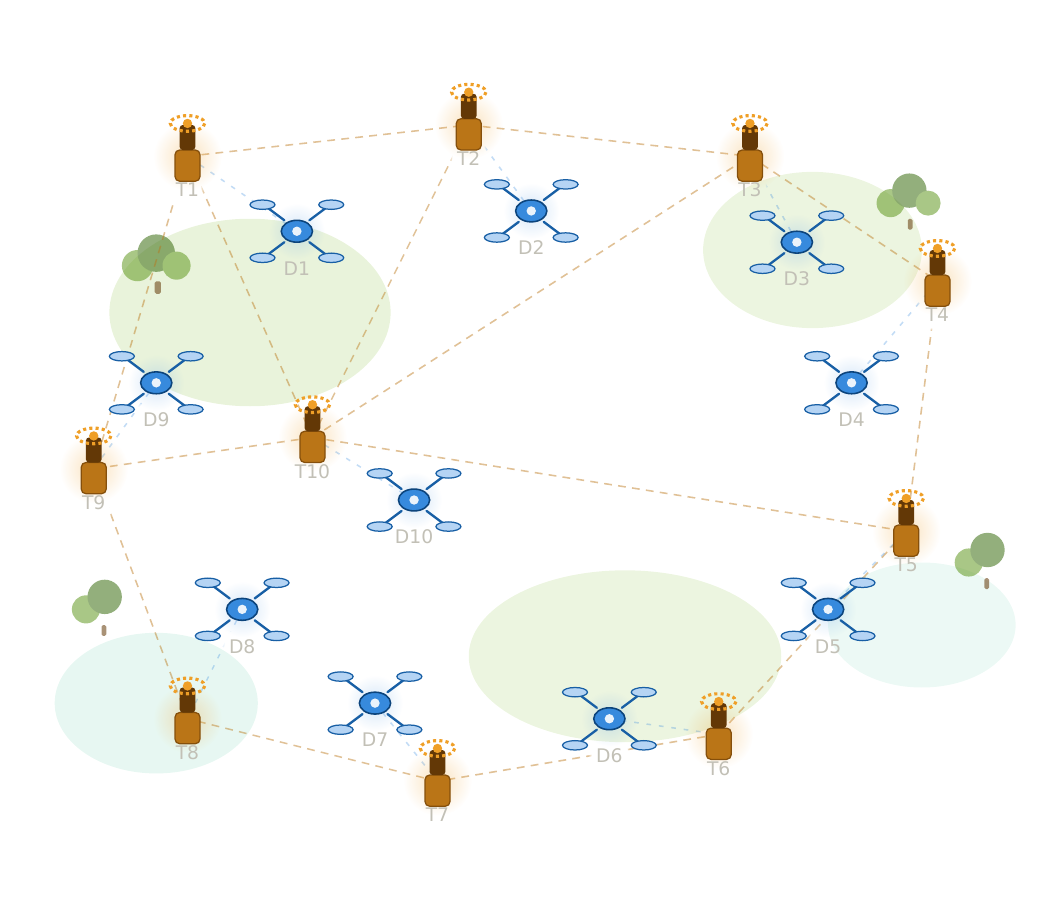}
        \caption{Multi-Robot}
        \label{fig:multirobot}
    \end{subfigure}%
    \begin{subfigure}[t]{0.17\textwidth}
        \centering
        \includegraphics[
            width=\textwidth,
            height=0.135\textheight,
            keepaspectratio,
            trim=15 25 15 20,
            clip
        ]{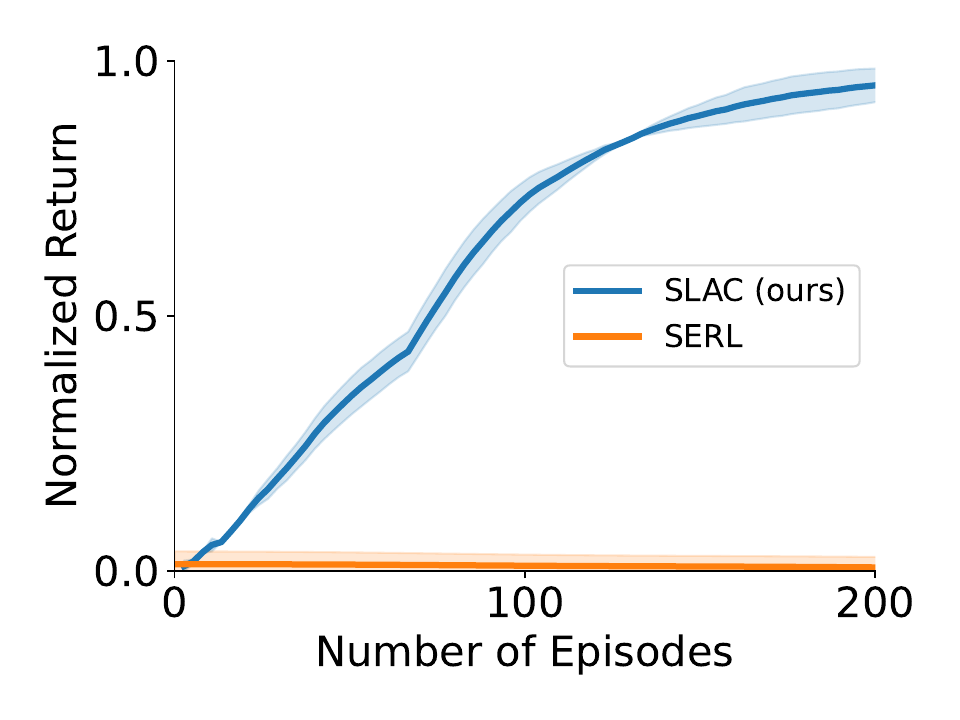}
        \caption{Parallel match}
    \end{subfigure}%
    \begin{subfigure}[t]{0.17\textwidth}
        \centering
        \includegraphics[
            width=\textwidth,
            height=0.135\textheight,
            keepaspectratio,
            trim=15 25 15 20,
            clip
        ]{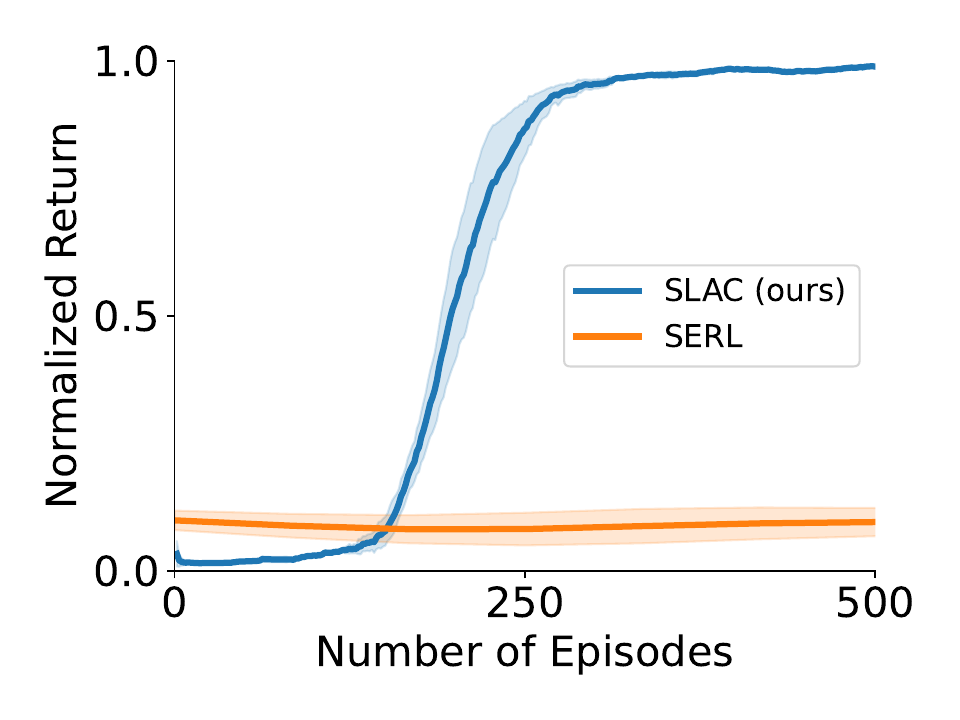}
        \caption{Sequential match}
    \end{subfigure}

    \vspace{-0.1cm}

    \caption{\methodname{} on the Multi-Robot Domain. SLAC is a general framework that can be applied to different domains. Results are averaged over 3 random seeds, with the shade showing the standard error. SLAC significantly outperforms the SERL baseline.}
    \label{fig:mr_rslt}
\end{figure}

Since \methodname{} is an embodiment-agnostic framework that does not require domain knowledge, we can in principle apply \methodname{} to any robots and even beyond robotics. We illustrate the broad applicability of \methodname{} on a simulated Multi-Robot domain~\citep{terry2021pettingzoo, hu2024disentangled}, where a centralized controller needs to simultaneously control 10 virtual drones to interact with different stations (Fig.~\ref{fig:multirobot}).
We consider two challenging downstream tasks, \emph{Parallel Match} and \emph{Sequential Match}, and describe them in detail in Appendix~\ref{app:mrenv}. We compare \methodname{} against SERL, and report the results in Fig.~\ref{fig:mr_rslt}. 
We can see that while SERL completely fails to learn these challenging tasks, \methodname{} is able to take advantage of the learned latent action and successfully solve these tasks. 
The strong performance in this domain suggests that SLAC's  structured action abstraction and decomposed learning objective are not specific to the Tiago domain, and can scale to high-dimensional coordinated control.


\subsection{Ablations}
\label{ss:abla}

Finally, we conduct ablation studies to examine how each component of \methodname{} contributed to the overall performances. 
\subsubsection{Ablations of Downstream Learning}
\label{ss:ds_abla}
We first ablate the effectiveness of each component of SLAC in downstream task learning. Specifically, we conduct ablation studies comparing against the following variations of \methodname{}:
\begin{itemize}[topsep=0pt,leftmargin=*]
\setlength{\itemsep}{0pt}
\setlength{\parskip}{0pt}
\setlength{\parsep}{0pt}
    \item \textbf{No Disentanglement}: where we remove the disentangled constraint during latent action space learning. As a result, our latent action space is no longer factored, and we can no longer apply Q-Function Decomposition during downstream learning since now all reward terms depend on the entire latent action vector.
    \item \textbf{On-Policy}: where we replace our proposed FLA-SAC with PPO~\citep{schulman2017proximal}, a state-of-the-art on-policy RL algorithm that has achieved many successes in Sim2Real RL. 
    \item \textbf{Not Temporally Extended}: where the task policy makes decisions at the same frequency as the latent action decoder (i.e. 10hz).
\end{itemize}

\begin{figure*}[t]
    \centering
    \begin{subfigure}[t]{0.18\textwidth}
        \centering
        \includegraphics[width=\textwidth, trim=15 25 15 20, clip]{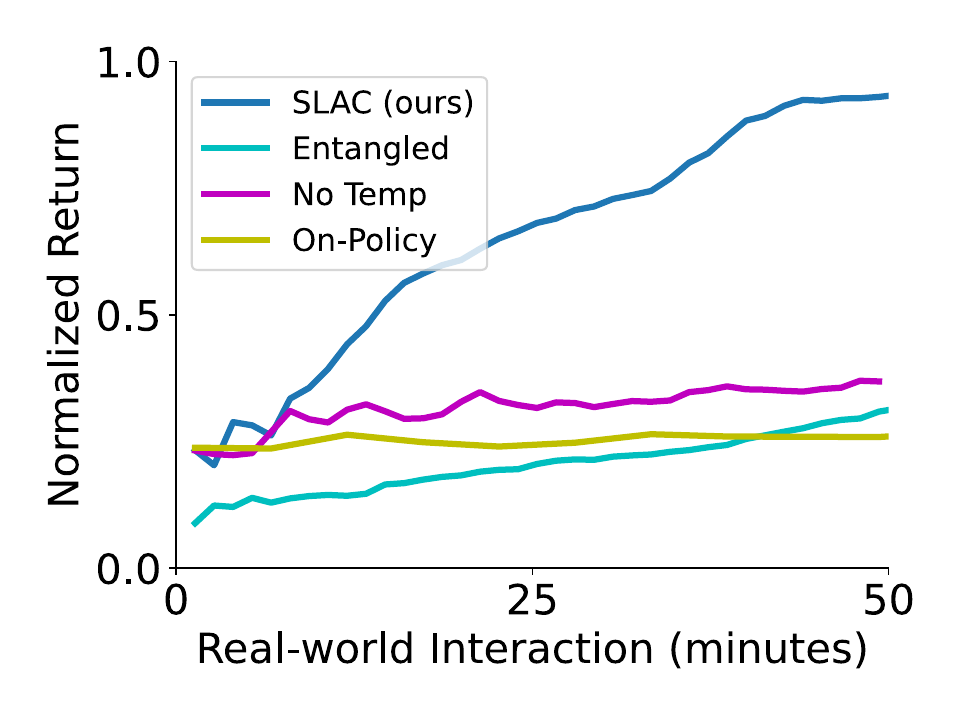}
        \caption{Board Wipe}
    \end{subfigure}%
    \hspace{0.015\textwidth}%
    \begin{subfigure}[t]{0.18\textwidth}
        \centering
        \includegraphics[width=\textwidth, trim=15 25 15 20, clip]{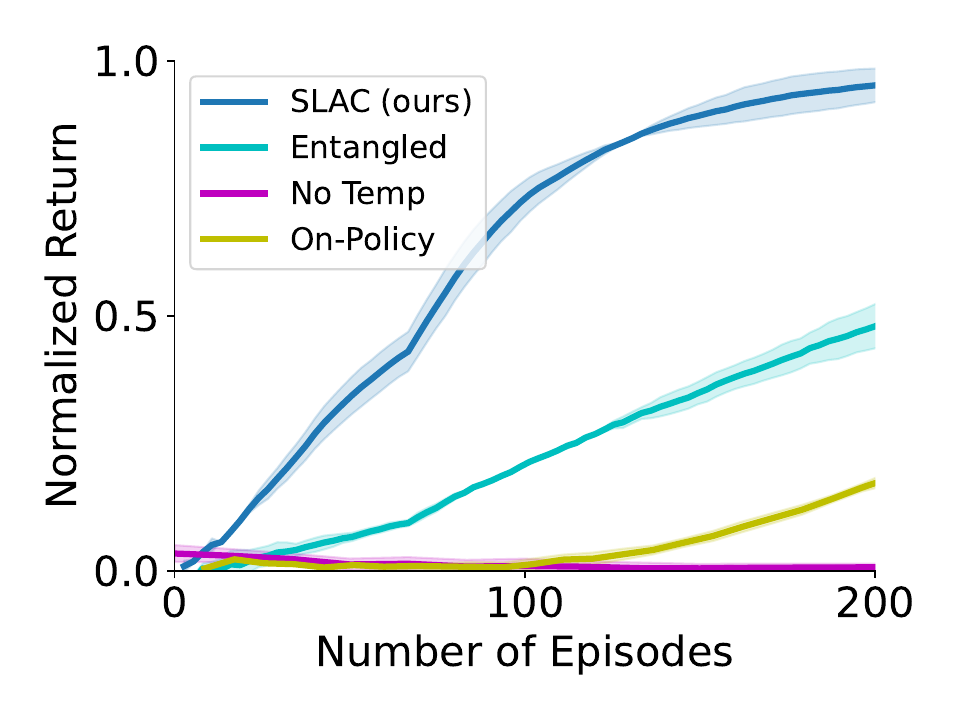}
        \caption{Parallel Match}
    \end{subfigure}%
    \hspace{0.015\textwidth}%
    \begin{subfigure}[t]{0.18\textwidth}
        \centering
        \includegraphics[width=\textwidth, trim=15 25 15 20, clip]{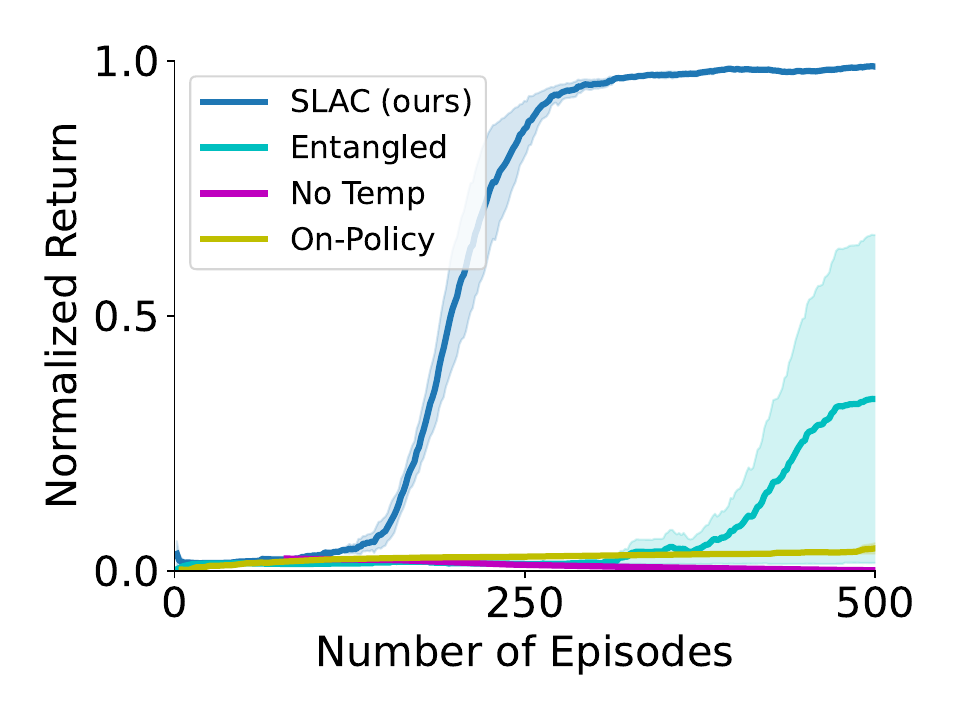}
        \caption{Sequential Match}
    \end{subfigure}%
    \hspace{0.015\textwidth}%
    \begin{subfigure}[t]{0.18\textwidth}
        \centering
        \includegraphics[width=\textwidth, trim=15 25 15 20, clip]{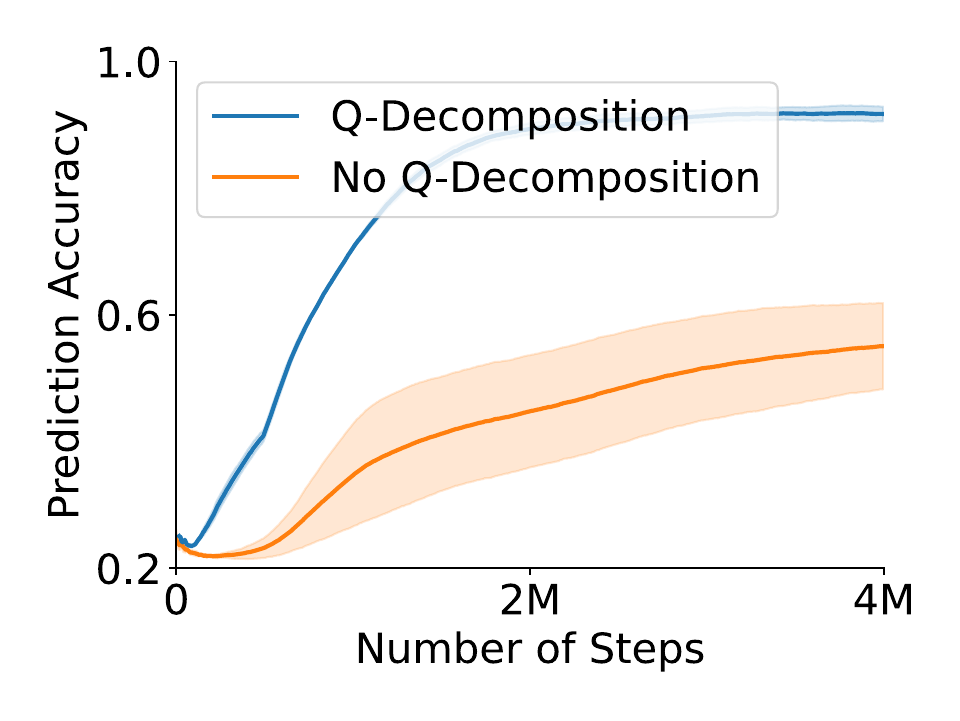}
        \caption{Q-dec: Multi-Robot}
    \end{subfigure}%
    \hspace{0.015\textwidth}%
    \begin{subfigure}[t]{0.18\textwidth}
        \centering
        \includegraphics[width=\textwidth, trim=15 25 15 20, clip]{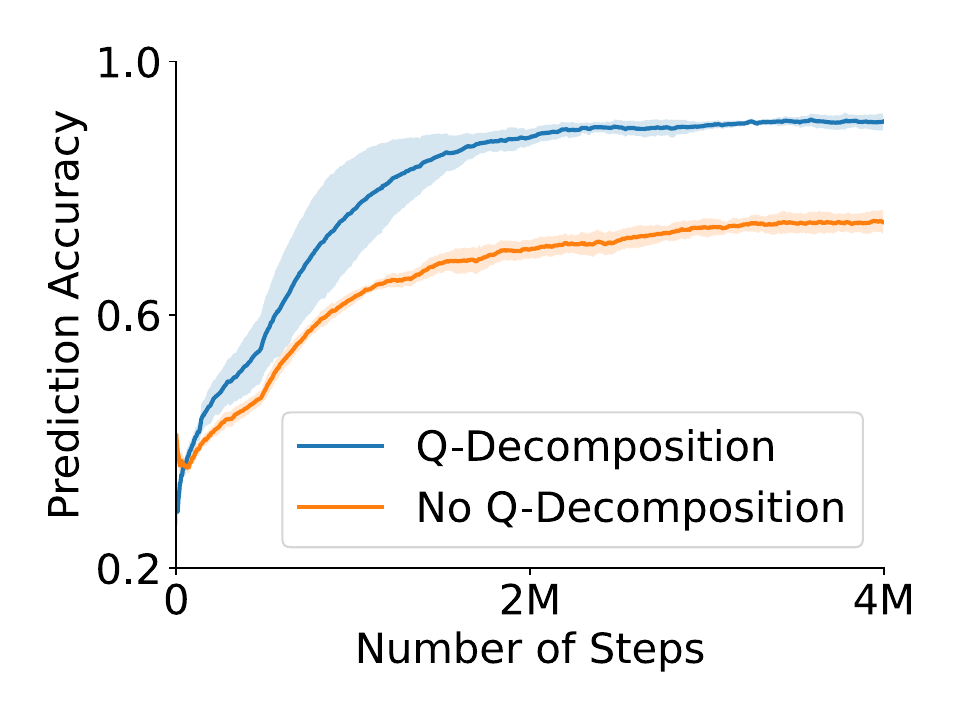}
        \caption{Q-dec: Tiago}
    \end{subfigure}
    \caption{
Ablation analyses for \methodname{}.
The first three plots evaluate the contribution of key techniques in \methodname{} to downstream task learning, showing that removing any component substantially reduces sample efficiency.
The last two plots evaluate the effect of Q-decomposition in latent action learning by measuring prediction accuracy ($\uparrow$) of latent action dimension $z^i$ from state factor $s^i$, averaged across 3 independent runs with different random seeds.
Higher prediction accuracy indicates that Q-decomposition encourages the policy to control state factors in more distinguishable ways, enabling more effective downstream learning.
}
    \label{fig:ablation}
\end{figure*}

We report the training curve for each of these variants of \methodname{} on both the mobile manipulation and the multi-robot domain in Fig.~\ref{fig:ablation}. We can see that removing any single component of \methodname{} results in a significant decrease in the learning efficiency, which shows that all components of SLAC downstream learning contribute to its performance.

 


\subsubsection{Ablations of Action Space Learning}
\label{ss:as_abla}
We examine the effect of Q-Decomposition introduced in Sec.~\ref{ss:dec} on latent action space learning. Specifically, we measure the classification accuracy of the latent action discriminators $q^i_\phi (z^i| s^i)$, averaged over all latent action dimensions, which indicates progress towards discovering diverse and distinguishable behaviors, with higher accuracy being better.
We depict our results in Fig.~\ref{fig:ablation}.
We observe that Q-decomposition significantly outperforms a monolithic Q network, suggesting that Q-decomposition is necessary for efficient learning of the latent action space.


\section{Related Work}


In this section, we review previous work in Sim-to-Real RL and real-world RL, the two dominant paradigms for applying RL to physical robots. Since the primary application in our experiments is mobile manipulation, we further discuss additional prior work in whole-body mobile manipulation.

\subsection{Sim-to-Real Reinforcement Learning}
While Reinforcement Learning (RL) provides a way for agents to learn sophisticated behaviors from trial and error, popular algorithms like PPO \citep{schulman2017proximal} are quite sample-inefficient and can require billions of samples before they converge. Furthermore, unsafe behaviors often occur during exploration. Many works have therefore resorted to performing the RL training completely in simulation \citep{hu2024flare,he2025asap,honerkamp2022n,yang2023harmonic,yokoyama2023adaptive,ma2023learning,fu2023deep,jauhri2022robot,Hu-RSS-23,li2025robotmover}, and zero-shot transfer the learned policy into the real-world. Such a procedure requires the simulation to have very high fidelity, and can pose significant challenges for simulated object creation, especially for tasks that are contact-rich/non-rigid \citep{tang2024deep}.
Unlike these works, \methodname{} relies on simulated interactions only to provide a suitable action space for downstream real-world RL, which reduces or eliminates the reliance on high-fidelity simulation.

\subsection{Reinforcement Learning in the Real World}
Directly doing RL in the real world offers a promising direction to avoid the requirement of high-fidelity simulation \citep{yang2024robot,smith2022walk,gupta2021reset,luo2024serl,julian2020scaling,zhang2024extractefficientpolicylearning,yin2025rapidlyadaptingpoliciesreal}.
However, these methods often target simple domains such as fixed-base manipulator, and fall short when applied to more complex embodiments such as whole-body mobile manipulation due to the high requirements for sample efficiency and safe exploration.
In the rare exceptions where a mobile manipulator does learn through trial-and-error in the real world~\citep{herzog2023deep,xiong2024adaptive,sun2022fully,mendonca2024continuously}, domain knowledge is often injected to simultaneously facilitate safety and efficient exploration, in the form of ad-hoc hand-crafted motion priors \citep{xiong2024adaptive, sun2022fully, mendonca2024continuously} and/or demonstrations \citep{xiong2024adaptive, herzog2023deep}.
By comparison, \methodname{} enables high-degree-of-freedom mobile manipulators to learn downstream tasks in the real world without relying on any demonstrations or hand-crafted behavior priors.

\subsection{Whole-body Mobile Manipulation}

Finally, since our main experiments involve whole-body mobile manipulation tasks, we review alternative methods that have been used to tackle this problem.

\textbf{Classical Motion Planning and Control:} A classic way to enable robots to perform tasks is through motion planning and control. When it comes to mobile manipulation, however, uncertainty and inaccuracy in localization frequently impede the accurate execution of planned trajectories~\citep{stilman2010global,dai2014whole,burget2013whole, wolfe2010combined, kalakrishnan2011stomp}. Moreover, when the robot needs to consider multiple objectives brought by whole-body motion, creating a motion planner is even harder, as it requires solving complex multi-objective optimization problems~\citep{huang2000coordinated,ratliff2009chomp,van2011lqg}. 
On the side of control, existing methods~\citep{seraji1998unified,yamamoto1992coordinating,sentis2006whole,dietrich2012reactive,papadopoulos2000planning, haviland2022holistic, pankert2020perceptive} resort to sophisticated prioritized solutions that require extensive tuning and pre-determined task priorities. 
Moreover, these methods assume accurate models of the robot and the environment, which often break in unstructured environments and with high-dimensional sensor signals (e.g. images).
By comparison, \methodname{} can autonomously learn closed-loop policies only based on onboard sensors, and does not require prior domain knowledge.


\textbf{Learning from Demonstrations:}
Recently, learning from demonstration has gained popularity as a powerful paradigm for learning robot behaviors, particularly for tabletop manipulation~\citep{mandlekar2021matters,chi2023diffusion,o2024open}.
As robot systems get more and more complex, however, collecting high-quality data can quickly become challenging due to high-degree-of-freedom embodiments that require coordinated control.
Even with carefully designed systems that only work for very specific embodiments~\citep{dass2024telemoma,jiang2025brs,fu2024mobile,li2024okami}, getting enough data for imitation learning remains hard and costly, especially for dynamic and contact-rich tasks.
\methodname{} does not require \textbf{any} demonstrations, and potentially can be applied to learn a wide range of tasks through autonomous interactions with the environment.

\section{Conclusion}
\label{sec:conclusion}

This article introduced \methodname{}, a framework that enables robots to learn policies safely and efficiently in the real world, by leveraging a latent action space trained in a low-fidelity simulation.
\methodname{} learns this latent action space through unsupervised skill discovery, and employs a novel sample-efficient RL algorithm to learn task policy in the \methodname{} latent action space. 
By treating the simulator as a proxy rather than an exact replicate of the physical world, \methodname{} reduces dependencies on high-fidelity simulations which are costly to create.
Evaluated on a set of contact-rich whole-body manipulation tasks, \methodname{} is able to solve the tasks in under an hour of real-world interaction, where baseline methods failed.
We believe \methodname{} provides a strong foundation for scaling real-world robot learning to increasingly complex and diverse tasks and embodiments.

Despite its strong empirical performance, \methodname{} is not without limitations. First, \methodname{} introduces an implicit trade-off related to the granularity of the latent action space.
For example, an identity mapping between the robot's raw action space and latent action space would make the downstream task policy capable of learning any task within the original capability of the robot, but would significantly reduce the sample efficiency (as shown in our results in Sec.~\ref{exp:moma}). Similarly, the temporal length of each latent action entails another tradeoff, where shorter latent actions give more control to the task policy at the expense of longer task horizons, which may hamper learning (as shown in our ablation studies in Sec.~\ref{ss:abla}). It is likely that the optimality of the latent action space will be strongly task-dependent.
Second, in the current \methodname{} framework, the latent action decoder is kept fixed during downstream learning. However, for more fine-grained tasks, we might benefit from finetuning the latent action decoder while training the task policy (e.g. via the option framework~\citep{bacon2017option}). 
Finally, while this article primarily focuses on the algorithmic side of real-world learning, we expect future engineering efforts in the automation of task reset and downstream reward generation (e.g., via VLM/LLM or a learned reward function) to further boost the downstream learning efficiency.

\begin{acks}
We thank members of RobIn and LARG for their valuable feedback on real robot setup (Shivin Dass, Arpit Bahety, Rutav Shah)  and the manuscript (Shivin Dass, Chris Huang, Yifeng Zhu). This work is supported in part by NSF
(FAIN-2019844, NRT-2125858), ONR (W911NF-25-1-0065), ARO
(W911NF-23-2-0004), DARPA (Cooperative Agreement HR00112520004 on Ad
Hoc Teamwork) Lockheed Martin, and UT Austin's Good Systems grand
challenge. Jiaheng Hu is funded in part by a PhD Fellowship from Two Sigma Investments, LP. 
Any opinions, findings, and conclusions or recommendations expressed in this material are those of the authors
and do not necessarily reflect the views of Two Sigma Investments, LP.
Peter Stone serves as the Chief Scientist of Sony AI and
receives financial compensation for that role.  The terms of this
arrangement have been reviewed and approved by the University of Texas
at Austin in accordance with its policy on objectivity in research.
\end{acks}


\bibliographystyle{SageH}
\bibliography{example.bib}

\clearpage
\appendix
\section{Appendix}
\label{s:appendix}

\subsection{Policy and Training Videos}
We encourage the reader to visit our \hyperlink{https://robo-rl.github.io/}{paper website} for videos of the \methodname{} training process in the real world and the learned policies.

\subsection{Properties of the Learned Latent Action Space}
\label{app:la}
Here, we discuss in detail the properties of the learned SLAC latent action space.
In short, the latent action space of SLAC is \textbf{environment-aware but task-agnostic}. 
It is task-agnostic because it is trained without a task reward, and is only encouraged to induce diversity in behavior following the USD objective. Therefore, \textbf{the same latent action space can tackle different tasks within a particular environment} (e.g., the “push to tray” and “sweep to bag” tasks in our experiment utilize the same latent action space). 
On the other hand, the latent action space is environment-aware because it is trained to induce diverse behavior \textit{in a particular scene} in simulation. Note that our latent action space is robust to small variations in the environment: for example, the action space learned in the board environment can be used to learn policies that wipe marks and avoid the trash can, even though there is neither a trash can nor wipeable marks in the simulated board environment.

\subsection{Discrete vs Continuous Latent Actions} 
\label{app:discrete}
The SLAC framework supports both discrete and continuous latent action. In our experiments, we made a deliberate choice to use a discrete latent action space since a discrete latent action space encodes a \textbf{compact} set of distinguishable behaviors, making it more amenable to hierarchical downstream RL. This is also a standard design choice of previous unsupervised skill discovery methods~\citep{eysenbach2018diversity,hu2024disentangled,park2023controllabilityawareunsupervisedskilldiscovery}. 

To apply \methodname{} to continuous latent actions, we can simply define each latent action component $z^i$ as a k-dimensional continuous vector (and therefore the total dimensions of the latent action space would be $k\times N$, where N is the total number of latent action components). Now we can define the prior distribution for each latent action component distribution $p(z^i)$ as an k-dimensional continuous uniform distribution (e.g. $\mathcal{U}[-1, 1]$). Notice that our objective remains unchanged, as the MI is well-defined no matter whether the latent action is discrete or continuous. Lastly, we need to change the output head of our latent action prediction networks $q$, such that instead of outputting a categorical distribution, it will output a continuous probability distribution (e.g. a multivariate Gaussian distribution with diagonal covariance).


\subsection{Universal Safety Reward}
\label{app:safe_r}
In \methodname{}, we employ a universal safety reward for ensuring that the learned latent action space is safe. This reward is the same across all our environments since it generally encourages safe and robust robot behaviors, and is defined as follows:

\begin{equation}
\begin{split}
r_\mathit{safe} &= -\lambda_1 \| a \|^2 - \lambda_2 \| a - a_{\text{prev}} \|^2 \\
&\quad - \lambda_3 \cdot \mathbb{I}_{\text{collision}} - \lambda_4 \cdot \mathbb{I}_{\text{force}}
\end{split}
\end{equation}

In all our experiments, we set $\lambda_1=0.01$, $\lambda_2=0.1$, $\lambda_3 = 0.2$, and $\lambda_4=0.05$.

\subsection{Hyperparameter}
Here, we present the hyperparameter for both the latent action decoder training and the downstream task learning. The same hyperparameters are shared across all tasks. 
We use a low-level step size $steps\_per\_skill = 50$ for all our experiments.

\begin{table}[h!]
\centering
\caption{Hyperparameters of Latent Action Decoder Learning. }
\begin{small}
\begin{tabular}{cccccc}
\toprule
& \multicolumn{2}{c}{\textbf{Name}}                         & \multicolumn{2}{c}{\textbf{Value}}     \\
\midrule
\multirow{10}{*}{SAC}
& \multicolumn{2}{c}{optimizer}                             & \multicolumn{2}{c}{Adam} \\
& \multicolumn{2}{c}{activation functions}                  & \multicolumn{2}{c}{ReLU}     \\
& \multicolumn{2}{c}{learning rate}                         & \multicolumn{2}{c}{$1 \times 10^{-4}$} \\
& \multicolumn{2}{c}{batch size}                            & \multicolumn{2}{c}{1024} \\
& \multicolumn{2}{c}{critic target $\tau$}                            & \multicolumn{2}{c}{0.01} \\
& \multicolumn{2}{c}{MLP size}                              & \multicolumn{2}{c}{[1024, 1024]} \\

& \multicolumn{2}{c}{n updates}                               &\multicolumn{2}{c}{2}  \\
& \multicolumn{2}{c}{\# of environments}                    & 
\multicolumn{2}{c}{16} \\
& \multicolumn{2}{c}{entropy coefficient $\alpha$}                    & \multicolumn{2}{c}{0.0} \\
& \multicolumn{2}{c}{log std bounds}                    & \multicolumn{2}{c}{[-10, 2]} \\
& \multicolumn{2}{c}{warmup samples}                    & \multicolumn{2}{c}{24000} \\
& \multicolumn{2}{c}{latent action dimension}                    & \multicolumn{2}{c}{$4^5$} \\
\bottomrule
\end{tabular}
\end{small}
\vspace{-0pt}
\label{tab:skill_train_params}
\end{table}

\begin{table}[h]
\centering
\caption{Hyperparameters of Downstream Learning. }
\begin{small}
\begin{tabular}{cccccc}
\toprule
& \multicolumn{2}{c}{\textbf{Name}}                         & \multicolumn{2}{c}{\textbf{Value}}     \\
\midrule
\multirow{10}{*}{FLA-SAC}
& \multicolumn{2}{c}{optimizer}                             & \multicolumn{2}{c}{Adam} \\
& \multicolumn{2}{c}{activation functions}                  & \multicolumn{2}{c}{ReLu}     \\
& \multicolumn{2}{c}{learning rate}                         & \multicolumn{2}{c}{$4 \times 10^{-4}$} \\
& \multicolumn{2}{c}{batch size}                            & \multicolumn{2}{c}{64} \\
& \multicolumn{2}{c}{critic target $\tau$}                            & \multicolumn{2}{c}{0.05} \\
& \multicolumn{2}{c}{MLP size}                              & \multicolumn{2}{c}{[256, 256]} \\

& \multicolumn{2}{c}{utd ratio}                               &\multicolumn{2}{c}{10}  \\
& \multicolumn{2}{c}{\# of environments}                    & 
\multicolumn{2}{c}{1} \\
& \multicolumn{2}{c}{entropy coefficient $\alpha$}                    & \multicolumn{2}{c}{0.1} \\
& \multicolumn{2}{c}{log std bounds}                    & \multicolumn{2}{c}{[-10, 2]} \\
& \multicolumn{2}{c}{warmup samples}                    & \multicolumn{2}{c}{60} \\
& \multicolumn{2}{c}{gumbel temperature}                    & \multicolumn{2}{c}{1.0} \\
\bottomrule
\end{tabular}
\end{small}
\vspace{-0pt}
\label{tab:policy_train_params}
\end{table}

\subsection{Mobile Manipulation Environment Description}
\label{app:env_moma}
In this section, we describe the two mobile manipulation environments that we tested \methodname{} on. In each environment, we apply our method to solve two different downstream tasks. We visualize the environments and the downstream tasks in Fig.~\ref{fig:env_viz}. In both environments, the robot has a 17-dimensional action space, corresponding to base velocity (3d), head camera joint position (2d), right end-effector delta pose (6d) and left end-effector delta pose (6d). The observation space of the task policy consists of a 320×240 RGBD image that is segmented and down-sampled to 50 points, and a 7-dimensional vector corresponding to the proprioceptive data. For all downstream tasks, we employ a relatively \textbf{sparse} reward that is only given at the end of a high-level policy step.  

\subsubsection{Board Environment}

\textbf{Simulation}
In the board environment, the robot is initialized in front of an interactable whiteboard. 
The decoder observation $o_\mathit{dec}$ consists of the proprioceptive data of the robot, the robot's previous action, the robot's distance and relative orientation with the board (which we estimate in the real world via a simple RANSAC line detector~\citep{fischler1981random}), and the end effector's contact history with the board. 
The latent action space is trained to maximize empowerment for the following state entities: board contact history, robot base position, robot view, and board contact force.

\textbf{Downstream Task 1: Clean Whiteboard}
The \textit{Clean Whiteboard} task requires the robot to identify the location of the board that needs to be wiped, and then use a sponge to clean up the identified region. Specifically, we define a composite task reward function with the following terms: 1) Encourage the robot to look at the target marker to wipe. 2) Encourage the successful removal of the target marker. 3) Encourage the robot to move towards the target marker. 4) Penalize large contact forces and any collision.
The task is considered successful if all four conditions are successfully achieved.

\textbf{Downstream Task 2: Wipe Board over Obstacles}
The \textit{Wipe over Obstacles} task is conceptually similar to the \textit{Clean Whiteboard} task, except that now there is an obstacle between the robot and the board. Thus, the robot needs to additionally learn to keep a reasonable distance from the obstacle and still be able to wipe the mark. The reward function is the same as \textit{Clean Whiteboard}.

\subsubsection{Table Environment}

\textbf{Simulation}
In the table environment, the robot is initialized in front of a table. The table is not interactive, but would incur a penalty if the robot collides with it.
The decoder observation $o_\mathit{dec}$ consists of the proprioceptive data of the robot, the robot's previous action, and the robot's distance and relative orientation with the table (which we again estimate in the real world via RANSAC~\citep{fischler1981random}). 
The latent action space is trained to maximize empowerment for the following state entities: robot left and right eef position relative to the table, robot base position, and robot view. 

\textbf{Downstream Task 3: Push Garbage on the Table to the Tray}
The \textit{Push to Tray} task requires the robot to push some garbage on the table into a tray that is also placed on the table. The reward function consists of the following terms: 1) Encourage the robot to look at the location of the garbage. 2) Encourage successfully pushing the garbage into the tray. 3) Encourage the robot to move towards the garbage. 4) Penalize large contact forces and any collision.

\textbf{Downstream Task 4: Sweep Garbage from the Table to the Bag}
For the \textit{Sweep to Bag} task, the robot is initialized with a bag in its left gripper. The goal of the task is to sweep the garbage into the bag, which requires the coordinated control of both robot arms and the base. The reward function consists of the following terms: 1) Encourage the robot to look at the location of the garbage. 2) Encourage successfully pushing the garbage off the table. 3) Encourage the robot to move its base towards the garbage. 4) Encourage the bag to be close to the garbage. 5) Penalize large contact forces and any collision.


\subsection{Multi-Robot Environment Description}
\label{app:mrenv}

The \emph{Multi-Robot} environment (Fig.~\ref{fig:multirobot}) consists of 10 robot drones that can move around, and 10 stations that the drone can interact with. For each robot, there is only one landmark that it can interact with.
The environment has a 70-dimensional observation space, consisting of the positions for each station and the positions and velocities for each robot. The action space is 50-dimensional, with 5 dimensions per robot that control their acceleration and interactions with the stations.

\textbf{Downstream Task 1: Parallel Match}
In this downstream task, all robots need to simultaneous decide whether to interact with their corresponding stations or not based on a sequence of binary indicators provided to the agents. Failing to follow the indicator will be penalized.

\textbf{Downstream Task 2: Sequential Match}
In this task, agents need to sequentially interact with their matching station following an instruction sequence given at the start of each episode. Interacting with stations in the wrong order will be penalized.

\end{document}